\documentclass{article}     

\usepackage{arxiv}
\usepackage[utf8]{inputenc} 
\usepackage[T1]{fontenc}    
\usepackage{hyperref}       
\usepackage{url}            
\usepackage{booktabs}       
\usepackage{amsfonts}       
\usepackage{nicefrac}       
\usepackage{microtype}      
\usepackage{lipsum}
\usepackage{graphicx}
\usepackage{caption}
\usepackage{subcaption}
\usepackage{multirow}
\usepackage{amsthm}
\usepackage{bm}
\usepackage{amsmath,amssymb,amsfonts}
\usepackage{algorithm}
\usepackage{algpseudocode}
\usepackage{wrapfig}
\title{On Flange-based 3D Hand-Eye Calibration for \\Soft Robotic Tactile Welding}
\author{
  Xudong Han\\
  Department of Mechanical and Energy Engineering\\
  Southern University of Science and Technology\\
  Shenzhen, China 518055\\
  \And
  Ning Guo\\
  Department of Mechanical and Energy Engineering\\
  Southern University of Science and Technology\\
  Shenzhen, China 518055\\
  \And
  Yu Jie\\
  Department of Mechanical and Energy Engineering\\
  Southern University of Science and Technology\\
  Shenzhen, China 518055\\
  \And
  He Wang\\
  School of Design\\
  Southern University of Science and Technology\\
  Shenzhen, China 518055\\
  \And
  Fang Wan$^{*}$\\
  School of Design\\
  Southern University of Science and Technology\\
  Shenzhen, China 518055\\
  \texttt{wanf@sustech.edu.cn}\\
  \And
  Chaoyang Song\thanks{Corresponding Author.}\\
  Design \& Learning Research Group\\
  Southern University of Science and Technology\\
  Shenzhen, China 518055\\
  \texttt{songcy@ieee.org}\\
}
\begin{document}
\maketitle
\begin{abstract}

    This paper investigates the direct application of standardized designs on the robot for conducting robot hand-eye calibration by employing 3D scanners with collaborative robots. The well-established geometric features of the robot flange are exploited by directly capturing its point cloud data. In particular, an iterative method is proposed to facilitate point cloud processing toward a refined calibration outcome. Several extensive experiments are conducted over a range of collaborative robots, including Universal Robots UR5 \& UR10 e-series, Franka Emika, and AUBO i5 using an industrial-grade 3D scanner Photoneo Phoxi S \& M and a commercial-grade 3D scanner Microsoft Azure Kinect DK. Experimental results show that translational and rotational errors converge efficiently to less than 0.28 mm and 0.25 degrees, respectively, achieving a hand-eye calibration accuracy as high as the camera's resolution, probing the hardware limit. A welding seam tracking system is presented, combining the flange-based calibration method with soft tactile sensing. The experiment results show that the system enables the robot to adjust its motion in real-time, ensuring consistent weld quality and paving the way for more efficient and adaptable manufacturing processes.
    
\end{abstract}
\newpage
\section{Introduction}
\label{sec:Intro}

    Depth sensor provides a versatile perception of the physical world with refined details through three-dimensional (3D) measurements. Since Microsoft's Kinect \cite{Sarbolandi2015KinectRange}, a wide range of consumer-grade 3D scanners has lowered the entry barriers when integrating robotic vision in research and applications \cite{Jarabo2017RecentAdvances}. Through optical perception, depth-sensing technologies translate the geometric details in the physical world into three-dimensional point cloud data concerning the camera frame \cite{Kolb2010ToF}. In areas including robotic welding \cite{Fan2019APrecision, Lu2023Automatic}, material handling \cite{Zhang2020ErrorCorrectable, Peng2021AHybrid}, and human-robot collaborations \cite{Shi2012LevelsOfHuman}, robotics researchers have shown a growing acceptance of adopting depth-sensing technologies \cite{Hyatt2019ConfigurationEstimation}, yet the robot-camera, or so-called hand-eye, calibration remains the first problem in practice \cite{Rout2019Advances, Wu2016Simultaneous, Wu2020HandEye}. 

    Industrial robots are usually built with excellent repeatability but relatively low accuracy, often requiring calibration using machine vision \cite{Huang2023DynamicParameter, Jiang2017KinematicAccuracy}. The repeatability problem is commonly solved by directly reading the sensor data saved at each joint \cite{Kluz2014TheRepeatability}. In contrast, the accuracy problem involves the inverse kinematic computation of a specific or target pose in the Cartesian space at the end-effector \cite{Zhang2021ASimultaneous}. The hand-eye calibration enhances the robot's accuracy by compensating the errors between the robot controller's computed pose and the camera's measured pose by the camera \cite{Kahn2014HandEye}.
    
    Hand-eye calibration is a 3D problem that can be solved using classical 2D cameras or by incorporating emerging 3D depth sensors. The classical problem of hand-eye calibration has been well-studied over the years \cite{Shah2012AnOverview}, which usually involves a robot as the \emph{robot}, a camera as the \emph{eye}, an end-effector as the \emph{hand}, and a high-precision calibration marker or object as the \emph{world} \cite{Wu2016Simultaneous}.
    
    \begin{itemize}
        \item The \textbf{hand-eye} calibration, by design, refers to the relationship between the Tool Center Point (TCP) and the ``eye'' camera. This is especially true in scenarios with a fixed end-effector as the hand to simplify the expression. However, in many academic and engineering applications, robotic researchers also choose to use the default TCP on the tool flange as the hand to enhance the re-usability of the calibration results. 
        \item The \textbf{tool-flange} calibration specifies the relationship between the default TCP at the tool flange and the actual TCP at the end-effector. The default TCP is directly accessible in most robot controllers, which the manufacturer already calibrates. When an end-effector is attached, one can directly refer to the technical data sheet for the tool-flange relationship. However, many end-effectors are customized according to the specific use, which may require further calibration to determine the tool-flange relationship.
        \item The \textbf{robot-robot} calibration refers to the case when multiple robots are used for collaborative tasks, such as dual-arm robots and robot-assisted surgeries. In practice, many multi-robot systems are designed with one robot attached with an ``eye'' camera and the other with an end-effector. The relative positioning between the robot base frames must be calibrated before use. 
    \end{itemize}
     
    The calibration problem can be mathematically formulated into two equations \cite{Wang2022Robot}. The first one is $AX=XB$, where $X$ is the unknown hand-eye transformation, $A$ involves the relative transformation of the robot's TCP, and $B$ involves the simultaneous relative transformation of the calibration object to the camera. The second one is $AX=YB$, where $X$ and we $Y$ are the unknown hand-eye and robot-world transformations, and $A$ and $B$ involve the poses of the robot's TCP and the calibration object, respectively. Solving these equations often requires numerical optimizations, and the accuracy is highly dependent on various sources of error, including mechanical errors, measurement noise, calibration fixture inaccuracies, algorithmic limitations, etc.
    
    The hand-eye calibration incorporating 3D sensors has also been extensively studied as industrial applications emerge. Recent work by \cite{Wu2020HandEye} proposed a 4D Procrustes Analysis Approach for the hand-eye calibration problem, where standardized objects are still required for implementation. \cite{Hu2013ARapid} proposed a hand-to-eye calibration method using a Bursa coordinate transform model through depth sensing. \cite{Kahn2014HandEye} designed a 3D calibration object with a curved surface such that its pose can be uniquely estimated using the iterative closest point (ICP) algorithm, demonstrating that 3D calibration provides more accurate results on average. \cite{Zhang2017StereoVision} addressed hand-eye calibration using a surgical robot with a stereo laparoscope by proposing a computationally efficient iterative method. \cite{Yang2018RoboticHandEye} adopted a sphere model as the calibration object and reformulated the hand-eye calibration problem to use only the calibration object's translation (3-DoF) data.

    Theoretically, the calibration object can be any object in the camera's view as long as the object's pose can be estimated \cite{Li2018Simultaneous}. However, there are a few limitations to the current solutions. In industrial applications such as assembly and manipulation, where high accuracy is required, standard calibration objects with high manufacturing precision are needed for hand-eye calibration, which is usually expensive \cite{An2016MethodFor}. In the eye-to-hand scenario, installing and removing the calibration object from the robot arm adds an extra burden to the already time-consuming deployment of robots \cite{kalia2019marker}. Moreover, involving an external calibration object brings another unknown robot-world transformation or increases the complexity of calibration equations \cite{Li2018Simultaneous}. With the growing market of robotic engineering, the standardization of robot design and manufacturing provides a rich set of geometric features that are directly measurable by the depth sensors \cite{mcgarry2022assessment}. The International Standard Organization (ISO) 9409-1:2004 defines the main dimensions, designations, and markings for a circular plate and a cylindrical shaft on the tool flange as the mechanical interface to ensure the exchangeability and orientation of end-effectors \cite{ISO9409-1}. There is a need for further research on the utilization of depth-sensing technologies to directly measure such standardized mechanical interfaces in 3D, which constitutes the focus of this paper. 

\begin{figure}[htbp]
        \centering
        \includegraphics[width=0.8\linewidth]{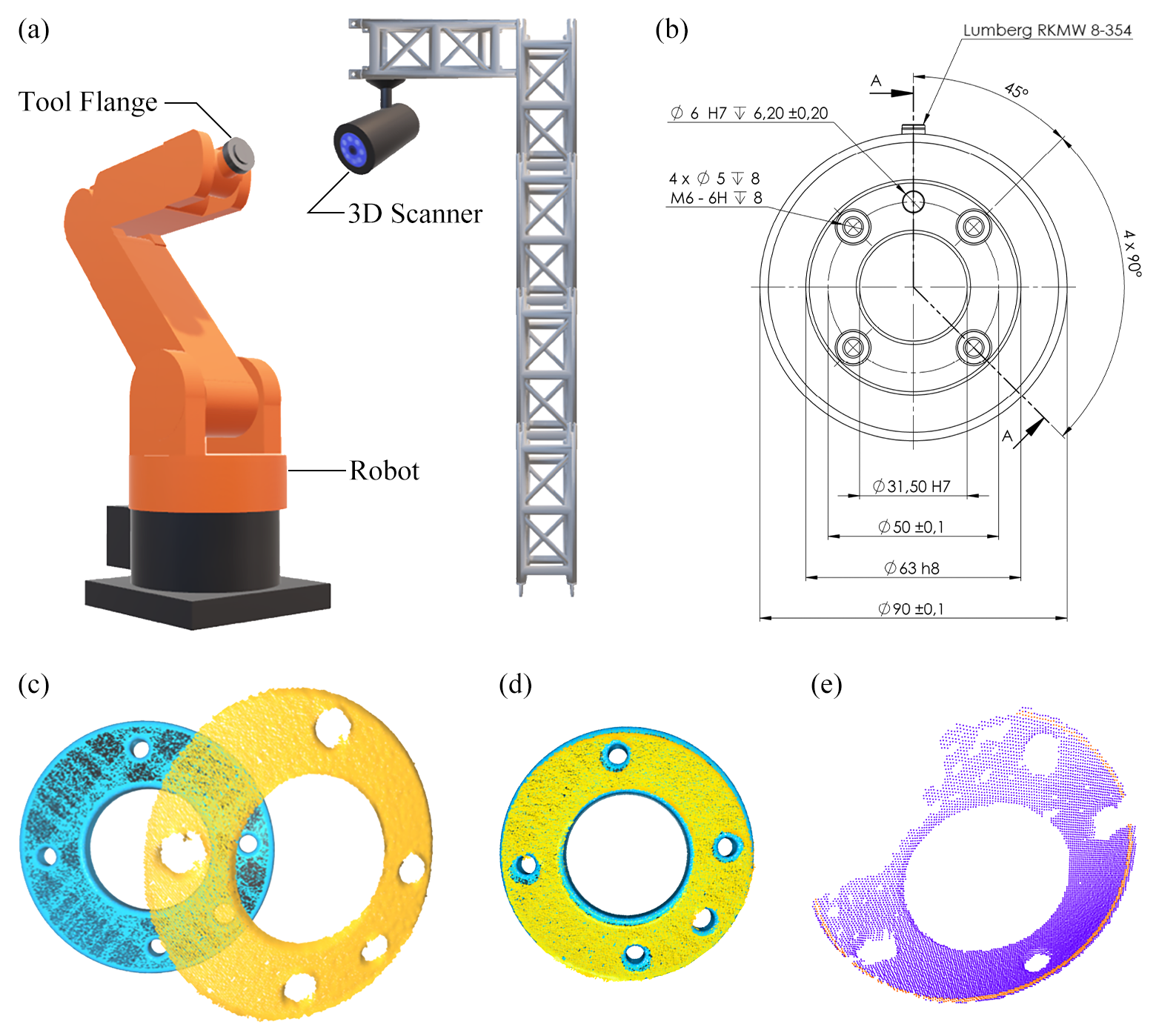}
        \caption{A direct hand-eye calibration based on a three-dimensional measurement of the standardized geometrical features on the robot's tool flange.}
        \label{fig:PaperOverview}
    \end{figure}

    In this paper, we propose a novel method using high-fidelity 3D scanners to directly measure the standardized geometric features on the tool flange of a robot for hand-eye calibration, as shown in Fig. \ref{fig:PaperOverview}. The setup of direct flange-based hand-eye calibration (Fig. \ref{fig:PaperOverview}a) includes a 3D scanner and a tool flange on a robot. Since the flange design of robots, such as UR5 (Fig. \ref{fig:PaperOverview}b), follows the ISO standards, the geometric features are readily identifiable. Fig. \ref{fig:PaperOverview}c shows the point clouds of the flange sampled from the CAD model (blue) and captured by the 3D scanner (yellow). With the proposed flange-based hand-eye calibration, the yellow point cloud can well match the blue point cloud, as shown in Fig. \ref{fig:PaperOverview}d. In addition, the proposed iterative method could also provide reliable circular feature calibration (orange) with partial point clouds (violet) (Fig. \ref{fig:PaperOverview}e). Unlike previous approaches that mainly measure calibration objects external to the robot system, the proposed method focuses on a calibration process using a 3D measurement of the geometric features within the robot system under international standardization. The proposed method effectively reduces system errors by removing unnecessary estimations and transformations during the calculation process of hand-eye calibration. 

    With the growing adoption of 3D scanners in robotic welding, the proposed method can potentially reduce the complexities in setting up the vision-based robot system for integration, as demonstrated in an original robotic welding system integrating 3D vision and soft tactile sensing (more details are in Section \ref{sec:softwelding}). Adopting high-precision 3D vision scanners in robotic welding enables robust and efficient weld seam tracking and fully autonomous welding \cite{Geng2024ANovel}. Accurate hand-eye calibration is the first step in the general process of seam tracking via a vision system \cite{Rout2019Advances, Geng2023ANovel}. Tactile sensors are also used when vision systems are not proficient, e.g., the weld seam is not entirely in the view of a camera or polluted environments \cite{Lei2020ATactual, Michael2020HapticBased}. The discrepancy between the robot's welding trajectory and the actual weld can be mitigated through the force feedback sensing provided by tactile sensors \cite{suwanratchatamanee2009robotic}. Owing to the outstanding flexibility and safety, tactile sensors fabricated from soft materials are extensively employed in robot tasks \cite{lepora2021soft, zhang2024evaluation}. \cite{Wu2024Vision} proposed a soft conical network structure with tactile sensing capability, and it is appropriate for fitting welding seams and providing force feedback. In this paper, the proposed calibration method, combined with the tactile sensor, is applied to the robotic welding to ensure consistent welding quality and adjust the motion in real-time.

    Contributions of this paper are listed as the following:
    \begin{itemize}
        \item Proposed a novel hand-eye calibration method by measuring the intrinsic design features of the robot system, i.e., tool flange and base mount, using high-fidelity 3D scanners;
        \item Implemented an iterative algorithm that effectively and efficiently optimizes the calibration accuracy as high as the camera's, probing the hardware limit;
        \item Conducted a quantitative and systematic evaluation of the calibration accuracy using several collaborative robots, industrial-grade, and consumer-grade 3D scanners;
        \item Presented a safe and adaptable welding seam tracking system that combines the proposed calibration method and soft tactile sensing.
    \end{itemize}
    
    The rest of this paper is structured as follows. Section \ref{sec:Methods} proposes the hand-eye calibration method via standardized robotic flanges and soft robot tactile welding system via multi-modal fusion. Section \ref{sec:Results} presents the simulation and experiment results using the flange-based calibration method and its application in robotic welding. Section \ref{sec:Discussion} discusses and evaluates the proposed method's effectiveness. Conclusion, limitations, and future works are summarized in Section \ref{sec:Conclusion}.

\section{Methods}
\label{sec:Methods}

\subsection{Hand-Eye Calibration via Standardized Robotic Flanges}

    The hand-eye calibration is a kinematic calibration problem, which usually involves four coordinate systems, including the base of the robot, the tool-mounting flange of the robot, the camera frame, and the calibration object frame, which are denoted by $\{\text{Base}\}$, $\{\text{Flan}\}$, $\{\text{Cam}\}$ and $\{\text{Mark}\}$ respectively. In this paper, we denote $_{B}^{A}H$ as the homogeneous transformation matrix of frame $B$ relative to $A$ and $\hat{H}$ (with hat) denotes unknown transformation to be calculated. The rest of the paper will focus mainly on the hand-eye calibration problem to demonstrate the proposed method, which can be further extended to other calibration configurations. The following notes are usually considered before analysis.
    
\begin{figure}[htbp]
        \centering
        \includegraphics[width=0.8\linewidth]{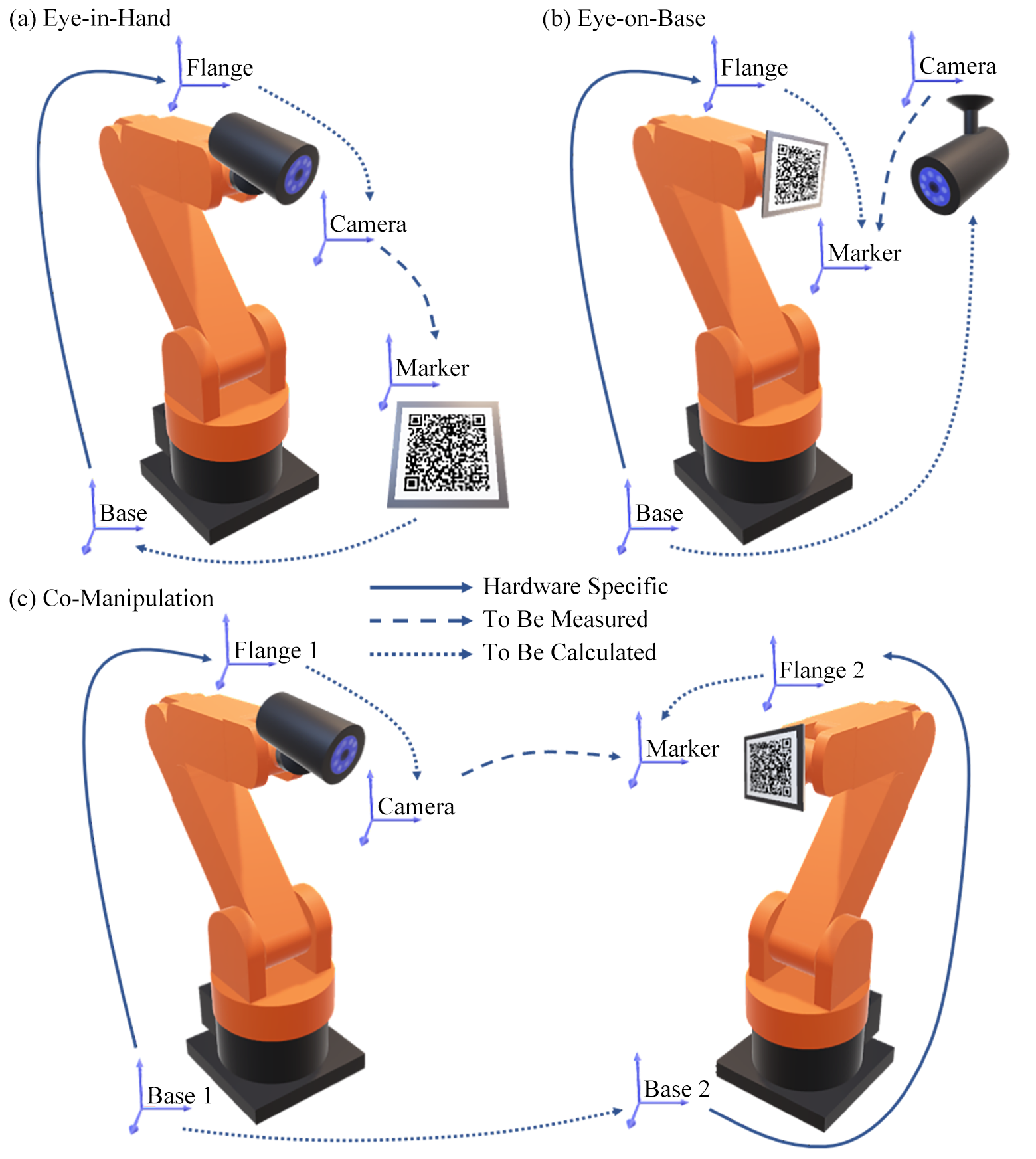}
        \caption{Common configurations of robot hand-eye calibration.}
        \label{fig:CommonHandEye}
    \end{figure}

    \begin{itemize}
        \item The transformation between a robot's $\{\text{Base}\}$ and $\{\text{Flan}\}$, i.e., $_{\text{Flan}}^{\text{Base}}H$, is usually known depending on the robot's specifications. 
        \item The transformation between the camera $\{\text{Cam}\}$ and object $\{\text{Mark}\}$, i.e., $_{\text{Mark}}^{\text{Cam}}H$, is a calculated matrix based on the camera's optical measurement of the object in the form of a 2D image or 3D point cloud.
        \item Depending on the object and camera placement relative to the robot, there are two common configurations of hand-eye calibration, namely Eye-in-Hand configuration (Fig. \ref{fig:CommonHandEye}a) and Eye-on-Base configuration (Fig. \ref{fig:CommonHandEye}b). 
        \item Another more advanced case of co-manipulation (Fig. \ref{fig:CommonHandEye}c) is when two robots are involved, equivalent to a combination of the Eye-in-Hand and Eye-on-Base. Furthermore, we will introduce another configuration of Eye-on-Arm calibration, where the camera can be fixed at any convenient location on the arm.
    \end{itemize}
    
\subsubsection{Four Configurations of Hand-Eye Calibration}

    \textbf{Eye-in-Hand Calibration}: For Eye-in-Hand configuration, the camera is mounted on the robot's wrist near the tool flange, which can be expressed as $_{\text{Cam}}^{\text{Flan}}\hat{H}$. On the other hand, the object is placed at a fixed location concerning the robot base, expressed as $_{\text{Base}}^{\text{Mark}}\hat{H}$. The \textit{hat} on top denotes that this transformation matrix will be calculated for calibration. Therefore, a closed-loop coordinate transformation can be formed as follows: 
    
    \begin{equation}
        _{\text{Flan}}^{\text{Base}}H\cdot{}_{\text{Cam}}^{\text{Flan}}\hat{H}\cdot{}_{\text{Mark}}^{\text{Cam}}H\cdot{}_{\text{Base}}^{\text{Mark}}\hat{H} = I.
        \label{eq:EyeInHand}
    \end{equation}

    Note that in Eq. \eqref{eq:EyeInHand}, $_{\text{Flan}}^{\text{Base}}H$ is a known matrix based on the robot's joint configuration, and $_{\text{Mark}}^{\text{Cam}}H$ is also a known one based on the camera's measurement. The iterative method is a standard solution to Eq. \eqref{eq:EyeInHand} by sampling multiple points within the camera's view range and the robot's dexterity space. For example, by moving the robot from point $p_{1}$ to $p_{2}$ in the configuration space, the calibration marker remains fixed to the robot base and the following two equations can be obtained concerning $\{\text{Flan}\}_{p_{1}}$ and $\{\text{Flan}\}_{p_{2}}$.
    
    \begin{equation}
        \begin{array}{c}
        _{\text{Flan}_{p_{1}}}^{\text{Base}}H\cdot{}_{\text{Cam}_{p_{1}}}^{\text{Flan}_{p_{1}}}\hat{H}\cdot{}_{\text{Mark}}^{\text{Cam}_{p_{1}}}H = _{\text{Flan}_{p_{2}}}^{\text{Base}}H\cdot{}_{\text{Cam}_{p_{2}}}^{\text{Flan}_{p_{2}}}\hat{H}\cdot{}_{\text{Mark}}^{\text{Cam}_{p_{2}}}H = {}_{\text{Mark}}^{\text{Base}}\hat{H}
        \end{array}, 
        \label{eq:EyeInHand-p1p2-1}
    \end{equation}
    \begin{equation}
        _{\text{Cam}_{p_{1}}}^{\text{Flan}_{p_{1}}}\hat{H} = {}_{\text{Cam}_{p_{2}}}^{\text{Flan}_{p_{2}}}\hat{H} = {}_{\text{Cam}}^{\text{Flan}}\hat{H}.
        \label{eq:EyeInHand-p1p2-2}
    \end{equation}

    By left multiplying and right multiplying both sides in Eq. \eqref{eq:EyeInHand-p1p2-1} by $_{\text{Flan}_{p_{2}}}^{\text{Base}}H^{-1}$ and $_{\text{Mark}}^{\text{Cam}_{p_{1}}}H^{-1}$ respectively, one can further rewrite Eq. \eqref{eq:EyeInHand-p1p2-1} as $AX=XB$ to solve for $_{\text{Cam}}^{\text{Flan}}\hat{H}$, where $X=_{\text{Cam}}^{\text{Flan}}\hat{H}$, $A=_{\text{Flan}_{p_{2}}}^{\text{Base}}H^{-1}\cdot{}_{\text{Flan}_{p_{1}}}^{\text{Base}}H$ and $B=_{\text{Mark}}^{\text{Cam}_{p_{2}}}H\cdot{}_{\text{Mark}}^{\text{Cam}_{p_{1}}}H^{-1}$. In practice, the points to be sampled can be as many as 10$\sim$30 points to improve the calibration accuracy.
    
    \textbf{Eye-on-Base Calibration}: In this paper, we use the term ``Eye-on-Base'' instead of ``Eye-to-Hand'' to differentiate it from the Eye-in-Hand configuration further, as the camera is usually mounted at a fixed location in the world frame concerning the robot's base frame, expressed as $_{\text{Cam}}^{\text{Base}}\hat{H}$. A high-precision calibration object is usually fixed on the robot's wrist near the tool flange, expressed as $_{\text{Mark}}^{\text{Flan}}\hat{H}$. For each robot pose, the following coordinate transformation establishes
    \begin{equation}
        _{\text{Flan}}^{\text{Base}}H\cdot{}_{\text{Mark}}^{\text{Flan}}\hat{H} = {}_{\text{Cam}}^{\text{Base}}\hat{H}\cdot{}_{\text{Mark}}^{\text{Cam}}H = {}_{\text{Mark}}^{\text{Base}}\hat{H},
        \label{eq:EyeOnBase}
    \end{equation}
    which can be symbolically expressed as $AX=YB$.
    
    Collaborative robots recently emerged as a viable solution for human-robot collaborative tasks such as co-manipulation, where multiple robots work collaboratively for an integrated task. Following the above notation, we can express the transformation of the co-manipulation as follows:
    \begin{equation}
        \begin{array}{c}
        _{\text{Flan1}}^{\text{Base1}}H\cdot{}_{\text{Cam}}^{\text{Flan1}}\hat{H}\cdot{}_{\text{Mark}}^{\text{Cam}}H = _{\text{Base2}}^{\text{Base1}}\hat{H}\cdot{}_{\text{Flan2}}^{\text{Base2}}H\cdot{}_{\text{Mark}}^{\text{Flan2}}\hat{H}
        \end{array}.
        \label{eq:CoManipulation}
    \end{equation}
    
    Solving Eq. \eqref{eq:CoManipulation} becomes more challenging as three unknown matrices need to be determined simultaneously. However, one can decompose the co-manipulation problem into a simultaneous calculation of an Eye-in-Hand problem and an Eye-on-Base one. The left side of Eq. \eqref{eq:CoManipulation} can be rewritten as an Eye-in-Hand problem for robot1 if the robot2 with a calibration object is at a fixed pose using Eq. \eqref{eq:CoMa-Breakdown}a. Similarly, the right side of Eq. \eqref{eq:CoManipulation} can be rewritten as an Eye-on-Base problem for robot2 when robot 1 with a camera is at a fixed pose in space using Eq. \eqref{eq:CoMa-Breakdown}b. As a result, Eq. \eqref{eq:CoManipulation} becomes the following.
    \begin{equation}
        \begin{cases}
        _{\text{Flan1}}^{\text{Base1}}H\cdot{}_{\text{Cam}}^{\text{Flan1}}\hat{H}\cdot{}_{\text{Mark}}^{\text{Cam}}H\cdot{}_{\text{Base1}}^{\text{Mark}}\hat{H} = I & (a)\\
        _{\text{Flan2}}^{\text{Base2}}H\cdot{}_{\text{Mark}}^{\text{Flan2}}\hat{H} = {}_{\text{Cam}}^{\text{Base2}}\hat{H}\cdot{}_{\text{Mark}}^{\text{Cam}}H & (b)\\
        _{\text{Base2}}^{\text{Base1}}\hat{H} = {}_{\text{Flan1}}^{\text{Base1}}H\cdot{}_{\text{Cam}}^{\text{Flan1}}\hat{H}\cdot{}_{\text{Cam}}^{\text{Base2}}\hat{H}^{-1} & (c)
        \end{cases}.
        \label{eq:CoMa-Breakdown}
    \end{equation}

    \textbf{Co-Manipulation Calibration}: Recent work by \cite{Wu2016Simultaneous} provides a comprehensive solution to the co-manipulation problem similar to Eq. \eqref{eq:CoManipulation}, in which the $_{\text{Cam}}^{\text{Flan1}}\hat{H}$ represents the Hand-Eye calibration problem for robot1; the $_{\text{Mark}}^{\text{Flan2}}\hat{H}$ represents the Tool-Flange calibration problem for robot2; and the $_{\text{Base2}}^{\text{Base1}}\hat{H}$ represents the Robot-Robot calibration problem between robot1 and robot2. These three problems can be integrated into a matrix equation of $AXB=YCZ$ for a simultaneous solution. Due to the complexity of the problem, the developed algorithm remains challenging to implement due to the uncertainties of the sensor noise \cite{Ma2018Probabilistic}.

\begin{figure}[H]
        \begin{centering}
        \textsf{\includegraphics[width=0.8\linewidth]{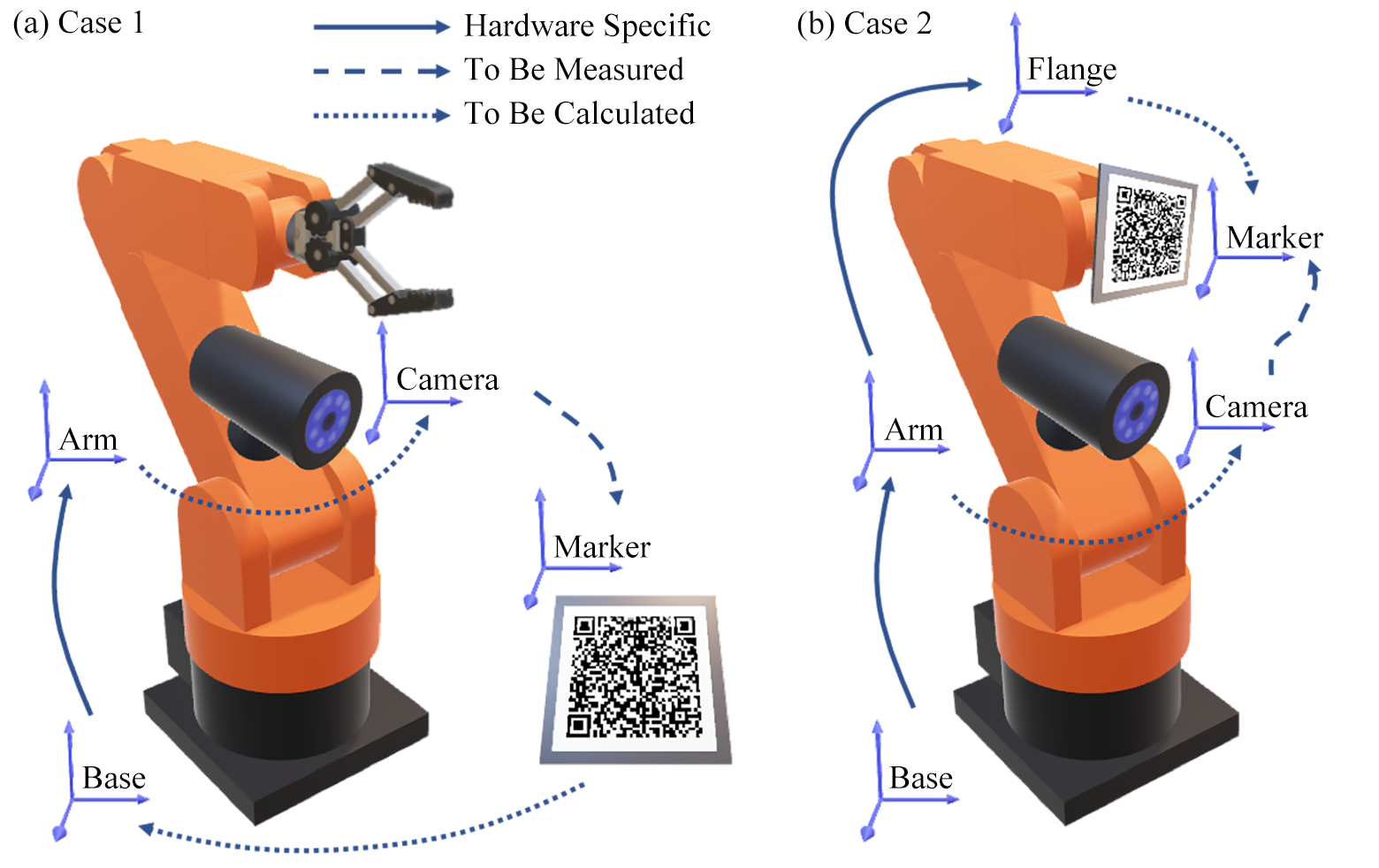}}
        \par\end{centering}
        \caption{Eye-on-Arm configuration for robot hand-eye calibration.}
        \label{fig:EyeOnArm}
    \end{figure}

    \textbf{Eye-on-Arm Calibration}: The relative placements of the camera and the marker differentiate the configurations of hand-eye calibration above. This naturally leads to the possibility of fixing the camera somewhere in the middle of the robot arm $\{\text{Arm}\}$, namely the Eye-on-Arm configuration in Fig. \ref{fig:EyeOnArm}. Depending on the placement of the calibration object, the first case of the Eye-on-Arm configuration is when the object is fixed on the robot's wrist near the robot flange, which is similar to the Eye-in-Hand configuration in Eq. \eqref{eq:EyeInHand} as 
    \begin{equation}
        _{\text{Arm}}^{\text{Base}}H\cdot{}_{\text{Cam}}^{\text{Arm}}\hat{H}\cdot{}_{\text{Mark}}^{\text{Cam}}H\cdot{}_{\text{Base}}^{\text{Mark}}\hat{H} = I.
    \end{equation}
    
    The second case is similar to the Eye-on-Base configuration, where the marker object is fixed to a point in space relative to the robot's base mounting flange. The coordinate transformation of this case can be written as 
    \begin{equation}
        _{\text{Flan}}^{\text{Base}}H\cdot{}_{\text{Mark}}^{\text{Flan}}\hat{H} = {}_{\text{Arm}}^{\text{Base}}H\cdot{}_{\text{Cam}}^{\text{Arm}}\hat{H}\cdot{}_{\text{Mark}}^{\text{Cam}}H,
    \end{equation}
    which is similar to Eq. \eqref{eq:EyeOnBase}. 
    
\subsubsection{Standardized Design of Robot Flanges}

\begin{figure}[htbp]
        \begin{centering}
        \textsf{\includegraphics[width=0.9\linewidth]{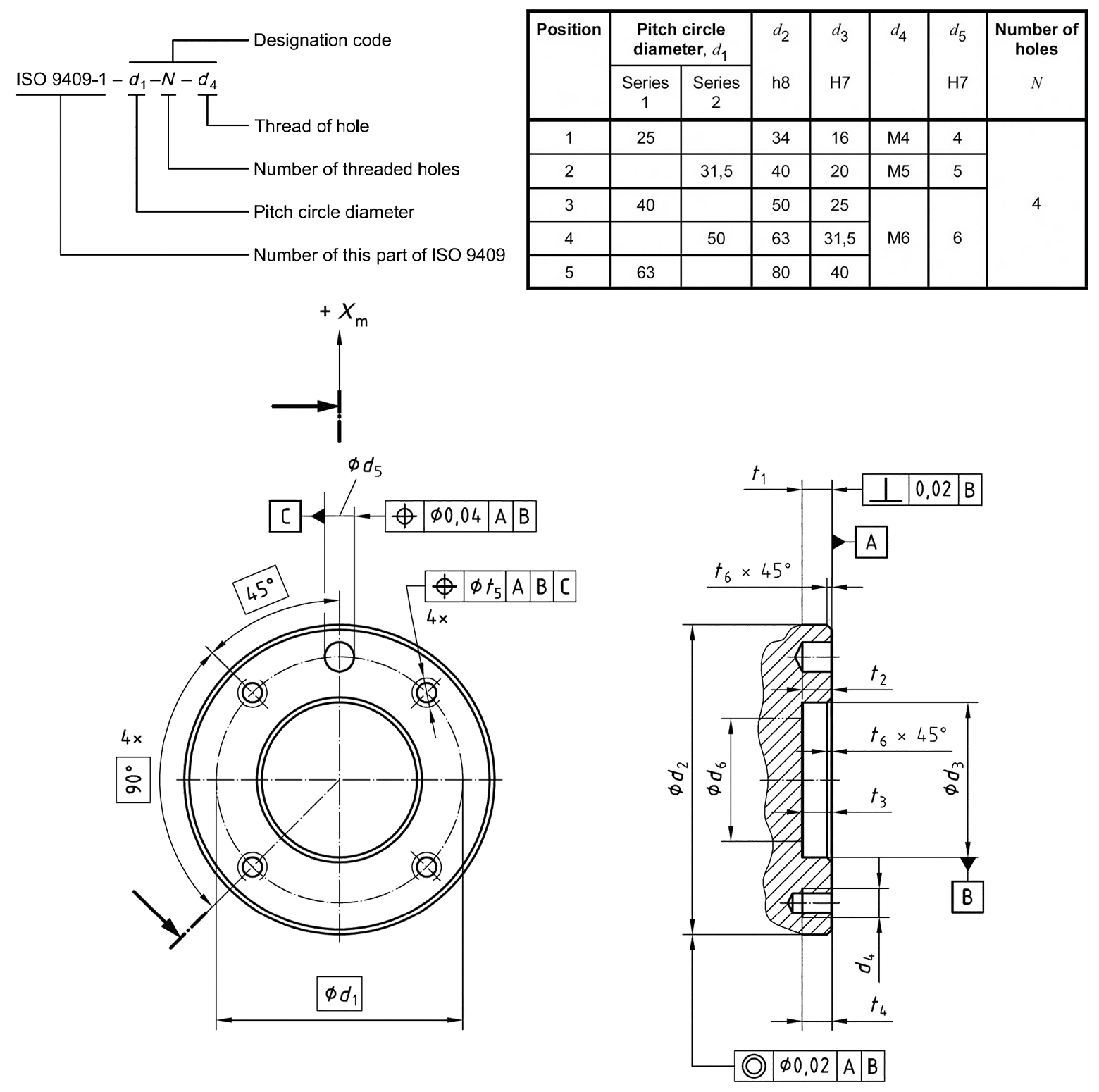}}
        \par\end{centering}
        \caption{The standardized geometric features on the flange of an industrial robot following ISO 9409-1: 2004 \cite{ISO9409-1}.}
        \label{fig:FlangeISO}
    \end{figure}

    The preliminary statistics of the World Robotics Report show a total of 3,903,633 units of operational stock of industrial robots worldwide in 2022, growing at an average of 13\% since 2017 \cite{Muller2023WorldRobotics}. Standardizing robot interfaces at various levels is critical to the reusability and exchangeability of robot systems, including mechanical, electrical, and communication. Among the International Standard Organization (ISO)'s catalog 25.040.30 industrial robots and robots, ISO 9409-1 specifies the design standardization of the mechanical interfaces or the fixture design on the tool flange \cite{ISO9409-1}. 
    Fig. \ref{fig:FlangeISO} is adapted from the latest version released in 2004, which specifies the critical mechanical interfaces, including the threaded holes referencing circle diameter in $d_{1}$, the flange's outer circle diameter in $d_{2}$, the size of the threaded holes $d_{4}$, the number of threaded holes $N$ to be used for fixture, etc. 
    
\begin{figure}[htbp]
        \begin{centering}
        \textsf{\includegraphics[width=0.95\linewidth]{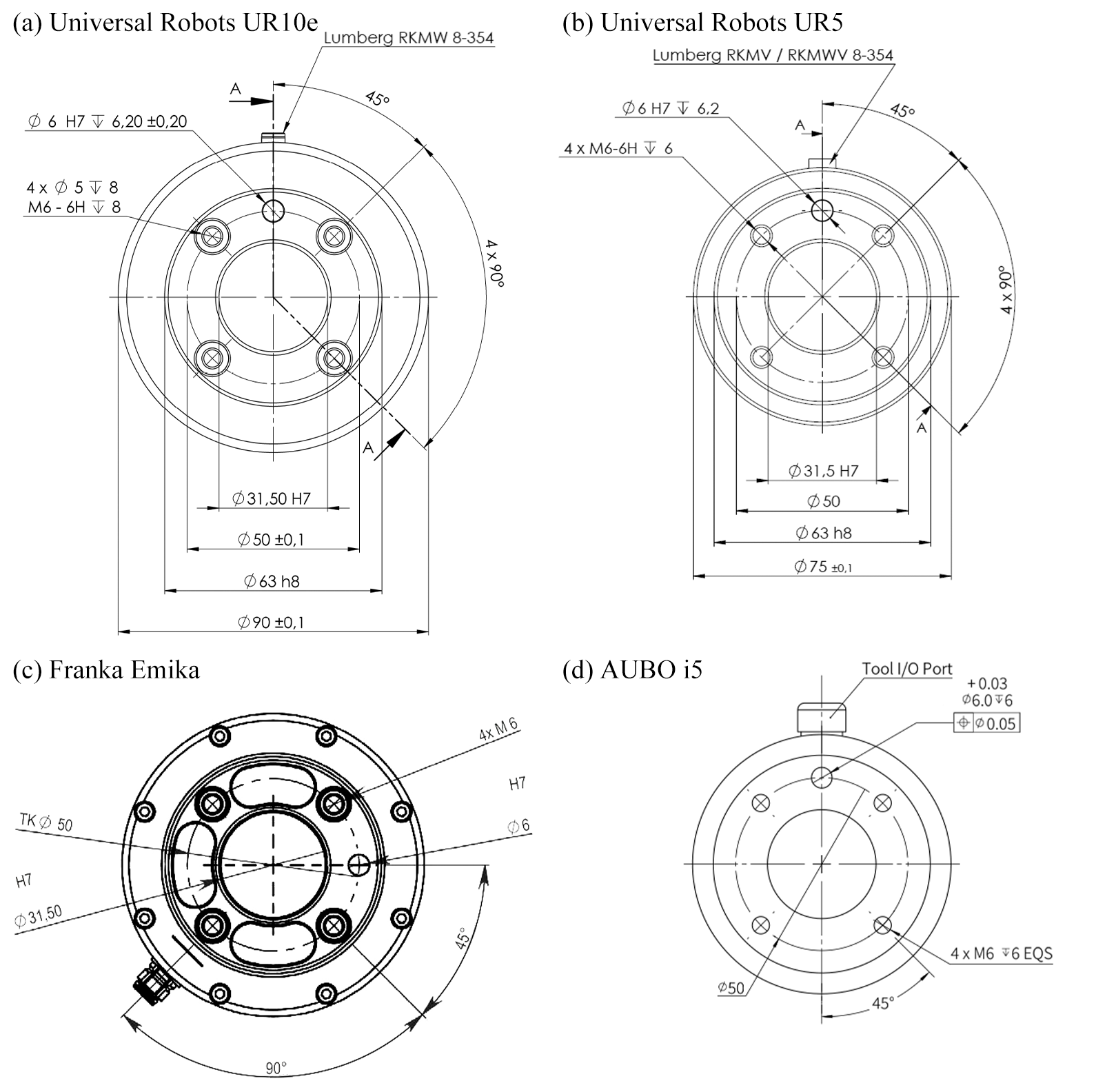}}
        \par\end{centering}
        \caption{The flanges of a few collaborative robots following ISO 9409-1-50-4-M6 standard: 
        (a) Universal Robots' UR10e \cite{UR10e}, 
        (b) Universal Robots' UR5 \cite{UR5}, 
        (c) Franka's Emika \cite{FrankaEmika}, and 
        (d) AUBO's i5 \cite{AUBOi5}.}
        \label{fig:CobotFlanges}
    \end{figure}

    A few flange design examples following the ISO 9409-1-50-4-M6 are reproduced in Fig. \ref{fig:CobotFlanges} among common collaborative robot brands. robot and end-effector manufacturers following the same designation code can be easily attached to accommodate different configurations of the robot systems for various applications. Such standardization also requires the manufacturers to meet specific manufacturing qualities to facilitate exchangeability. This study proposes to utilize such standardized design features for direct hand-eye calibration using high-fidelity 3D scanners, which will be explained next.

\subsubsection{Flange-based Hand-Eye Calibration}

    In this section, the Eye-on-Base configuration is used to demonstrate the proposed method of flange-based hand-eye calibration. To the authors' best knowledge, it is the first time that the tool-mounting flange of the robot is directly used as a calibration reference, especially when 3D depth sensing is used as the ``eye'' camera. As regulated by ISO 9409-1, the tool flanges are usually designed in a circular shape. Therefore, the center of the flange, which is also the robot's Tool Center Point (TCP), is selected as the referencing point for hand-eye calibration. As a result, the hand-eye calibration problem is reformulated as 
    \begin{equation}
        \left[\begin{array}{c}
        ^{\text{Base}}p_{i}\\
        1
        \end{array}\right]={}_{\text{Cam}}^{\text{Base}}H\cdot\left[\begin{array}{c}
        ^{\text{Cam}}p_{i}\\
        1
        \end{array}\right],
        \label{eq:FlangeEquation}
    \end{equation}
    
    where $\{ (^{\text{Base}}p_{i},{}^{\text{Cam}}p_{i})\mid i=1,2,...,n\} $ is a set of coordinate pair about the TCP relative to $\{\text{Base}\}$ and $\{\text{Cam}\}$, respectively. As the marker point is also the TCP of the robot arm, $^{\text{Base}}p_{i}$ can be directly obtained in the robot controller, whereas $^{\text{Cam}}p_{i}$ can be calculated by point cloud algorithms. Then, the hand-eye calibration problem becomes finding the least-squares estimation of transformation parameters between two sets of 3-DoF data as the following, $\min\frac{1}{n}\sum_{i=1}^{n}\left\Vert _{\text{Cam}}^{\text{Base}}R\cdot{}^{\text{Cam}}p_{i}+{}_{\text{Cam}}^{\text{Base}}t-{}^{\text{Base}}p_{i}\right\Vert$, which has an SVD-based optimal solution of $_{\text{Cam}}^{\text{Base}}\hat{H}$ as
    \begin{equation}
        \begin{cases}
        _{\text{Cam}}^{\text{Base}}\hat{R} = USV^{T} & (a)\\
        _{\text{Cam}}^{\text{Base}}\hat{t} = {}^{\text{Base}}\mu - c\cdot{}_{\text{Cam}}^{\text{Base}}\hat{R}\cdot{}^{\text{Cam}}\mu & (b)
        \end{cases},
        \label{eq:optimalH}
    \end{equation}
    where
    \begin{equation}
        \begin{cases}
        ^{\text{Base}}\mu = \frac{1}{n}\sum_{i=1}^{n}{^{\text{Base}}p_{i}}\\
        ^{\text{Cam}}\mu = \frac{1}{n}\sum_{i=1}^{n}{^{\text{Cam}}p_{i}}\\
        \sum = \frac{1}{n}\sum_{i=1}^{n}{(^{\text{Base}}p_{i} - {}^{\text{Base}}\mu)(^{\text{Cam}}p_{i} - {}^{\text{Cam}}\mu)^{T}}\\
        c = \frac{n\times \text{tr}(DS)}{\sum_{i=1}^{n}{\left\Vert ^{\text{Cam}}p_{i} - {}^{\text{Cam}}\mu\right\Vert ^{2}}}
        \end{cases},
        \label{eq:optimalH-2}
    \end{equation}
    and let the singular value decomposition of $\sum$ be $UDV^{T}$ and 
    \begin{equation}
        S = 
        \begin{cases}
        I & \textrm{if }\text{det}(\sum_{\text{cov}})\geq0\\
        \textrm{diag}(1,1,...,1,-1) & \textrm{if }\text{det}(\sum_{\text{cov}})<0
        \end{cases}.
        \label{eq:optimalH-1}
    \end{equation}

    At least four non-coplanar points are required to estimate a unique transformation matrix \cite{Wong2018OptimalLinear}. The optimal solution essentially represents the transformation matrix $_{\text{Cam}'}^{\text{Base}}\hat{H}$ from $\{\text{Base}\}$ to a calculated camera frame $\{\text{Cam}'\}$. 

\subsubsection{Solvability Analysis}

    During calibration, the robot arm is moved to poses such that the tool flange plane faces the 3D scanner. To obtain a high-quality point cloud of the tool flange, the angle between the normal vector of the tool flange and the optical axis should be less than a desirable threshold $\theta_{\text{max}}$. The position of the TCP relative to the 3D scanner is estimated in two steps. 
    \begin{itemize}
        \item First, pass-through and statistical filters are applied to the original point cloud of the scene to remove backgrounds such as table and floor and to remove noises. The cloud point of the flange plane can be, therefore, isolated using basic geometry segmentation algorithms in point cloud library \cite{Rusu20113DIsHere} together with geometric constraints, such as the segmented cluster can not have a range more extensive than the diameter of the tool flange. 
        \item Second, the center of the flange plane is estimated using the RANSAC algorithm \cite{Raguram2012USAC}. Moreover, we applied a model check so that only circles within the desired radius range would go to the verification stage, improving the search algorithm's efficiency.
    \end{itemize}

\begin{figure}[htbp]
        \begin{centering}
        \includegraphics[width=0.8\linewidth]{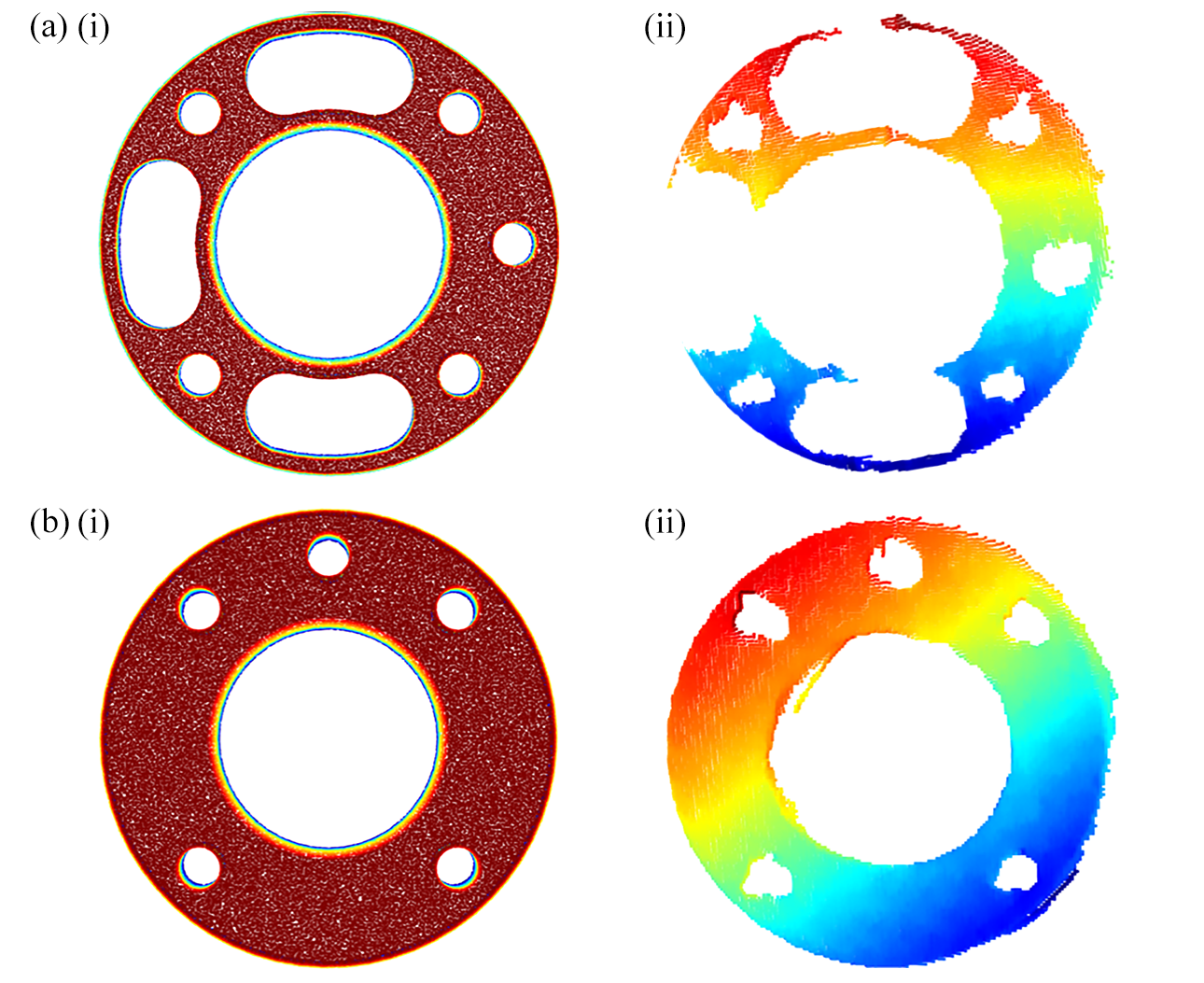}
        \par\end{centering}
        \caption{Point clouds of tool flanges on (a) Franka Emika and (b) UR5 from (i) CAD model and (ii) the respective point clouds measured by a 3D scanner.}
        \label{fig:GeneratedFlanges}
    \end{figure}

    While the re-projection error is usually adopted as the error metric for 2D hand-eye calibration, it does not apply to the proposed method in this paper as our method inherently includes the fitting error of the flange circle. It is easy to acquire the 3D model of the tool flange to generate a ground true point cloud in the tool flange coordinate $^{\text{Flan}}P_{\text{true}}$ in Fig. \ref{fig:GeneratedFlanges} in which the color is plotted according to the $z$-axis values of the point cloud.
    
    Using a known pose of the robot arm $_{\text{Flan}}^{\text{Base}}H_{v}$ for verification, we can obtain the ground true point cloud in the robot coordinate. Given the hand-eye transformation matrix, the corresponding measured point cloud $^{\text{Cam}}P_{v}$ can also be transformed into the robot coordinate. Then, one can align the ground true point cloud and the measured point cloud using the ICP algorithm \cite{Chen1992ObjectModelling}. Hence, we define the calibration error as
    \begin{equation}
        e_{\text{icp}} = 
        \begin{cases}
        \left\Vert \left[\begin{array}{c}
        ^{\text{Cam}}P_{v}\\
        1
        \end{array}\right]-{}_{\text{Base}}^{\text{Cam}'}\hat{H}\cdot{}_{\text{Flan}}^{\text{Base}}H_{v}\cdot\left[\begin{array}{c}
        ^{\text{Flan}}P_{\text{true}}\\
        1
        \end{array}\right]\right\Vert _{\text{icp}}\\
        +\infty,\textrm{if ICP fails}
        \end{cases},
        \label{eq:ErrorMetric}
    \end{equation}
    where $_{\text{Base}}^{\text{Cam}'}\hat{H}$ is the transformation matrix from calculated $\{\text{Cam}'\}$ to $\{\text{Base}\}$, $||^{\text{Cam}}P_{v}-^{\text{Cam}'}P_{v}||_{\text{icp}}$ calculate the rotation error $\delta R\in SO(3)$ and translation error $\delta t\in\mathbb{R}^{3}$ on the Euclidean group of rigid-body motions $SE(3)$ such that the two point clouds are registered. The ICP error metric $e_{\text{icp}}$ essentially registered the ground true camera frame $\{\text{Cam}\}$ and the calculated camera frame $\{\text{Cam}'\}$, namely $_{\text{Cam}}^{\text{Cam}'}H$, which is more appropriate and informative than traditional 2D error metric. In 2D hand-eye calibration, it is common to find that the hand-eye calibration error in position is small at the center of the camera's field of view and increases in marginal areas. The rotation error of the hand-eye calibration causes this phenomenon. With the help of a 3D scanner, the reason can be immediately verified and visualized. Before the calibration starts, the verification point cloud $P_{v}$ is suggested to be collected at a pose where the robot arm will mostly work around.
    
    A common problem during hand-eye calibration in 2D or 3D is the sampling quality, which usually requires further optimization. Typical issues include partial sampling, occlusion, and inaccurate circle fitting using the standardized RANSAC algorithm. Therefore, the analytical solution must collect more than four points to ensure a high-quality calibration. 

    Therefore, we propose an online iterative calibration method described in Algorithm \ref{alg:IterativeCalibration} to collect as few points as possible and increase the efficiency of the calibration process. The main difference is that it includes a self-verification mechanism such that the online calibration process becomes a closed loop. The goal is to maintain an optimized pool of four pairs of point cloud and robot pose to minimize the ICP error metric in $SE(3)$, which requires a real-valued cost metric $\Vert\cdot\Vert_{\text{cost}}:SE(3)\mapsto\mathbb{R}$. The calibration process keeps adding new data pairs to the pool and retaining the optimal four pairs with the least cost. The online calibration process stops once the cost metric has achieved a target error $e_{\text{required}}$. In practice, the design of the cost metric $\Vert\cdot\Vert_{\text{cost}}$ can be flexible according to the application scenario\footnote{Codes are available at \href{https://github.com/ancorasir/flange_handeye_calibration}{https://github.com/ancorasir/flange\_handeye\_calibration}.}. 
    \begin{algorithm}[htbp]
        \caption{The iterative flange calibration algorithm.}
        \label{alg:IterativeCalibration}
        \renewcommand{\algorithmicrequire}{\textbf{Input:}}
        \renewcommand{\algorithmicensure}{\textbf{Output:}}
        \begin{algorithmic}[1]
            \Require {Ground true point cloud $P_{\text{true}}$, verification point cloud $P_{v}$, flange radius $R$, calibration threshold $e_{\text{required}}$, maximum iteration $k<k_{\text{max}}$;}
            \Ensure {Hand-eye transformation $\hat{H}_{\text{optimal}}$ and ICP error metric $e_{\text{icp}}$}
            \State {$k\leftarrow0$;}
            \State {Collect a set of four initial pairs of point cloud and robot pose $S_{k}=\{^{\text{Cam}}P_{i}, ^{\text{Base}}\text{pose}_{i}\mid i=1,2,3,4\}$;}
            \State {Calculate $\hat{H}$ and $e_{\text{icp}}$ and $\hat{H}_{\text{optimal}}\leftarrow\hat{H}$, $e_{\text{optimal}}\leftarrow e_{\text{icp}}$;}
            \While {$\Vert e_{\text{optimal}}\Vert_{\text{cost}}>e_{\text{required}}$ \textbf{and} $k\leq k_{\text{max}}$}
                \State {Collect a new pair of point cloud and robot pose $S_{k}^{\prime}$;}
                \For {$i$ in $\{1,2,3,4\}$}
                    \State {Replace the $i$th pair in $S_{k}$ by the new pair of data;}
                    \State {Calculate $\hat{H}$ and ICP error $e_{\text{icp}}$;}
                    \If {$\Vert e_{\text{icp}}\Vert_{\text{cost}}\ensuremath{}\geq\ensuremath{}\Vert e_{\text{optimal}}\Vert_{\text{cost}}$}
                        \State {Undo the replacement}
                    \Else 
                        \State {$\hat{H}_{\text{optimal}}\leftarrow\hat{H}$}
                        \State {$e_{\text{optimal}}\leftarrow e_{\text{icp}}$}
                    \EndIf
                \EndFor
            \EndWhile
        \end{algorithmic}
    \end{algorithm}

\subsection{Soft Robotic Tactile Welding via Multi-Modal Fusion}
\label{sec:softwelding}
\subsubsection{Touch-based Welding Tool Design}

\begin{figure}[htbp]
        \centering
        \includegraphics[width=0.8\linewidth]{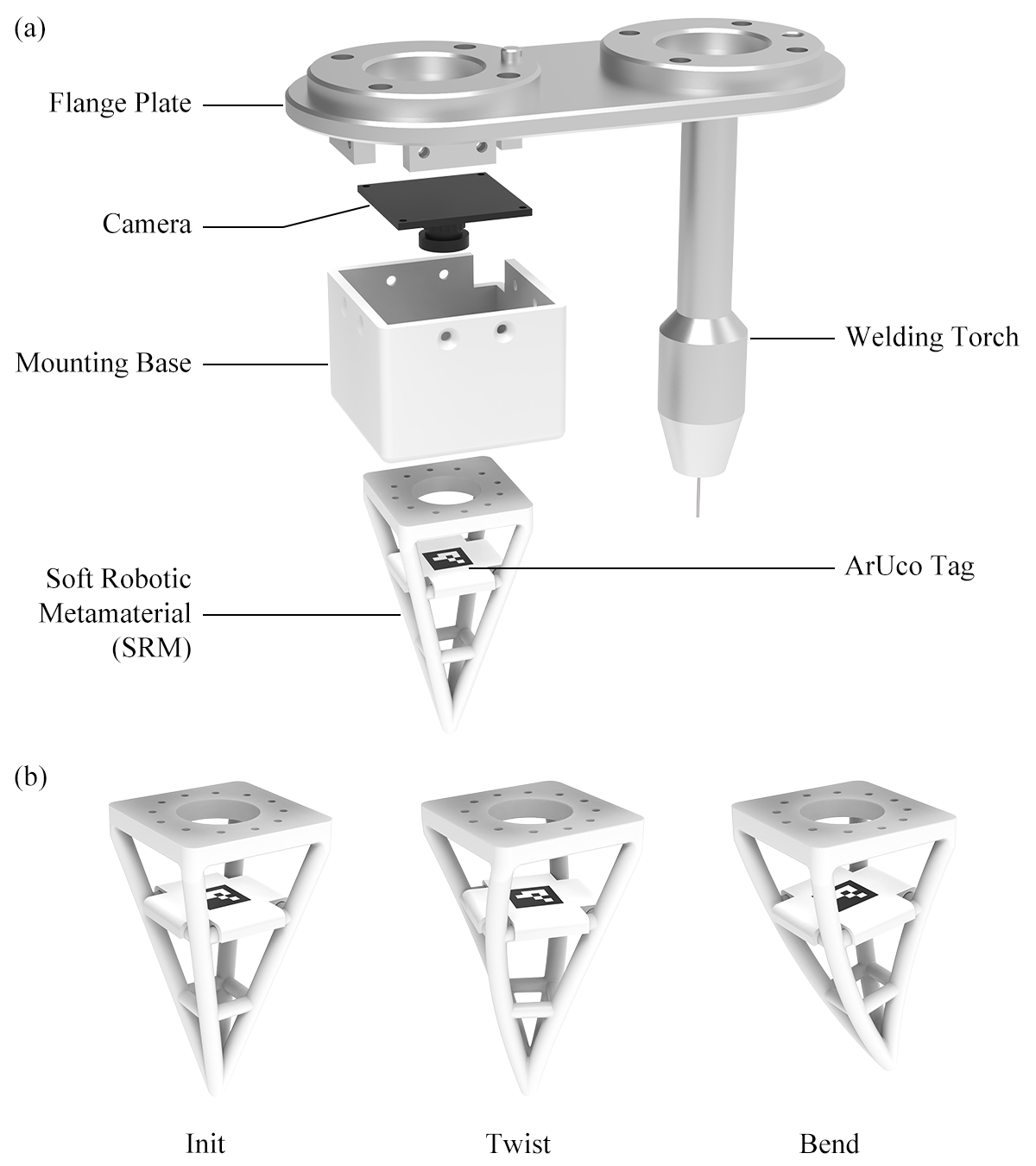}
        \caption{Welding tool with soft robotic metamaterial (SRM).}
        \label{fig:WeldingTool}
    \end{figure}

    We proposed a touch-based welding tool with soft robotic metamaterial (SRM) based on our previous work \cite{Wu2024Vision}. As shown in Fig. \ref{fig:WeldingTool}a, the welding tool consists of a flange plate, a camera, a mounting base, an ArUco tag, a soft robotic metamaterial (SRM), and a welding torch. The whole tool can be mounted on the robot with the left flange. The key features are the suitable flange-like geometric feature facing towards the 3D scanner on the top of the plate for hand-eye calibration and a soft robotic metamaterial (SRM) with tactile sensing capability. In Fig. \ref{fig:WeldingTool}b, The SRM has omni-directional deformation capability when contacting surrounding objects, such as welding seam, which is suitable for local seam path detection. During welding, the SRM deforms when there is a deviation between the robot and the intended paths. This deformation could be predicted by tracking the displacement and pose of the ArUco tag.

\subsubsection{Welding Seam Tracking by Soft Touch}\label{sec:welding_seam_tracking}

    Before welding starts, a welding seam path can be captured and extracted using 3D vision sensors. However, due to unavoidable error sources, such as welding seam feature extraction or hand-eye calibration, the welding path for the robot to execute can be unpredictable. It may lead to significant deviations from the intended path. Therefore, the fusion of tactile welding seam tracking with 3D vision could offer a comprehensive solution to address these challenges. 

    The SRM introduces a novel approach for tracking the welding seam by leveraging a simple mechanism of deformation servo. Commanding the SRM tip to maintain contact with one side of the welding seam effectively follows the planned welding path from 3D vision, thus compensating for potential noise and errors inherent in the vision data. Specifically, this is achieved by directing the soft torch to follow the planned path only in its tangential direction while maintaining a predefined deformation along its normal direction. This strategy ensures that the SRM consistently identifies and tracks the side of the welding seam without losing its position. Additionally, the tip of the SRM is continuously guided to remain within the V-shaped welding seam, allowing for the recording of its position and generating accurate real torch path points based on this recorded information. This innovative approach demonstrates a promising method for reliable welding seam tracking with the SRM, mitigating potential errors and ensuring precise and consistent weld quality.
    \begin{figure}[htbp]
        \centering
        \includegraphics[width=0.8\linewidth]{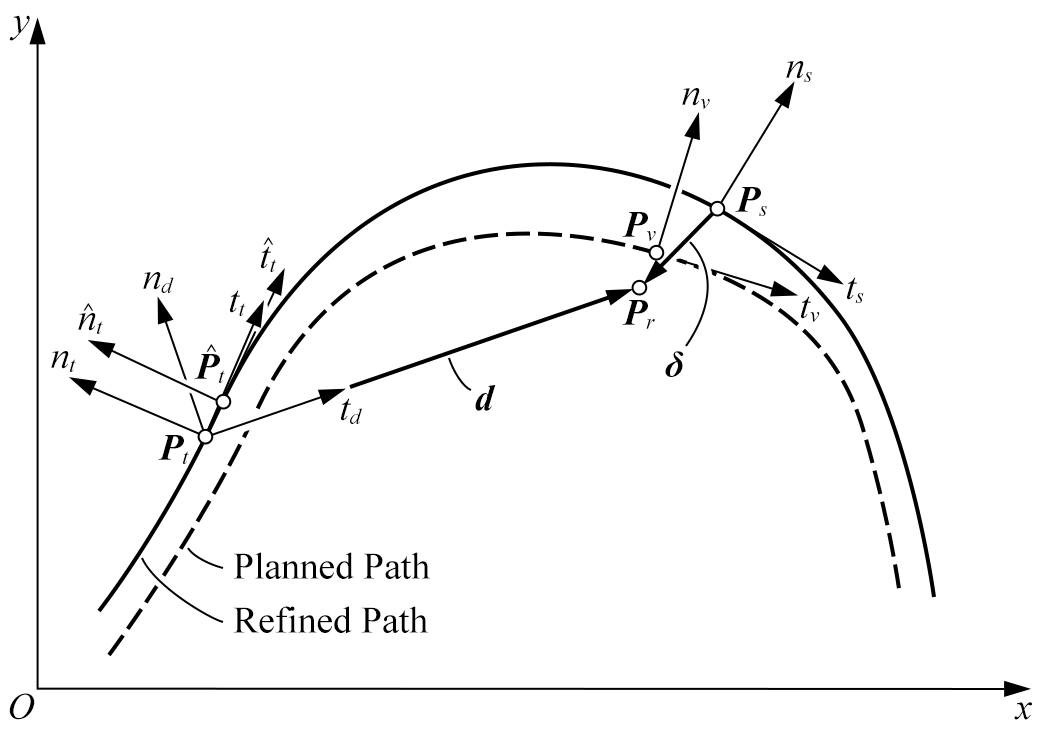}
        \caption{Online robotic welding trajectory generation with tactual weld seam tracking.}
        \label{fig:Welding_seam_tracking}
    \end{figure}
    
    Here, we describe the online robotic welding trajectory generation enhanced by tactual weld seam tracking. As shown in Fig. \ref{fig:Welding_seam_tracking}, we consider the online welding trajectory generation problem in a 2D plane. Denote $\mathbf{P}_{r} = [x_{r},y_{r}]$ as the position vector of the robot end effector in a fixed reference frame $\{O\}$, $\mathbf{P}_{s} = [x_{s},y_{s}]$ as the position vector of SRM tip in reference frame $\{O\}$, and $\mathbf{P}_{t}=[x_{t},y_{t}]$ as the position vector of welding torch in the same reference frame. 
    
    The direction of a touch-based welding tool can be described by orthonormal vectors obtained from the rotation angle $\alpha$ of the robot end effector. These vectors are defined as $\Vec{t}_{d} = [\cos{\alpha},\sin{\alpha}]$ and $\Vec{n}_{d} = [-\sin{\alpha},\cos{\alpha}]$, respectively. Because of the rigid connection between the position vector of the robot end effector and the welding torch, we can express this relationship using the following equation:
    \begin{equation}\label{eq:rigid_connection}
        \begin{cases}
            \mathbf{d} = \Vert \mathbf{d} \Vert  \Vec{t}_{d} \\
            \mathbf{P}_{r} = \mathbf{P}_{t} + \mathbf{d}  \\
        \end{cases},
    \end{equation}
    where, $\mathbf{d}$ represents the position difference vector between the robot end effector and the welding torch, and  $\mathbf{P}_{r}$ can be obtained from the robot controller.
    
    At each point along the welding path planned using 3D vision data, when the SRM makes contact with one side of the welding seam, the deformation can be sensed and calculated by utilizing the displacement of the Aruco marker $\bm{\delta}\in{\mathbb{R}^{2}}$ detected through the in-finger camera of the SRM. This computation is represented as:
    \begin{equation}\label{eq:path_point_correction}
        \begin{cases}
            \mathbf{P}_{r}^{t_s} = \mathbf{P}_{s}^{t_s} + \bm{\delta}^{t_s}  \\
            \mathbf{P}_{r}^{n_s} = \mathbf{P}_{s}^{n_s} + \bm{\delta}^{n_s}  \\
        \end{cases},
    \end{equation}
    where, $\mathbf{P}_{r}^{t_s}$, $\mathbf{P}_{s}^{t_s}$ and $\bm{\delta}^{t_s}$ represent the corresponding position vectors projected along the tangential direction at each path point, while  $\mathbf{P}_{r}^{n_s}$, $\mathbf{P}_{s}^{n_s}$ and $\bm{\delta}^{n_s}$ denote vectors along the normal direction.

    To uphold the SRM's deformation along the normal direction, we initially establish a desired deformation along the normal direction, denoted as $\bm{\delta}_{d}^{n_s}$. Subsequently, to maintain this directional deformation, the robot end effector velocity can be commanded using the following formula:
    \begin{equation}\label{eq:tcp_velocity_command}
        \begin{cases}
            \mathbf{V}_{r}^{t_s} = \mathbf{V}_{const}  \\
            \mathbf{V}_{r}^{n_s} = -k_{p}(\bm{\delta}^{n_s} - \bm{\delta}_{d}^{n_s})  \\
        \end{cases},
    \end{equation}
    where the robot end effector velocity along the tangential direction $\mathbf{V}_{r}^{t_s}$ maintains $\mathbf{V}_{const}$, which is a constant norm vector but changes direction. While in the normal direction, the velocity of the robot end effector $\mathbf{V}_{r}^{n_s}$ opposes the difference between the current detected Aruco marker displacement in the normal direction and the pre-defined deformation. This action serves as a resilience measure to maintain the deformation at the desired level, with the action being governed by the parameter $k_{p}$.
    
    During the deformation servo, the position of the SRM tip, denoted as $\mathbf{P}_{s}$, is recorded as a refined welding seam path to be subsequently executed by the real welding torch.  And this path will be executed by the real welding torch. As the welding torch is rigidly coupled to the robot end effector, the implicated velocity of the torch can be calculated as:
    \begin{equation}\label{eq:gun_implicated_velocity}
        \begin{cases}
            \mathbf{V}_{t}^{t_{d}} = \mathbf{V}_{r}^{t_{d}}  \\
            \mathbf{V}_{t}^{n_{d}} = \mathbf{V}_{r}^{n_{d}} + \omega \, \Vert \mathbf{d} \Vert\, \Vec{n}_{d} \\
        \end{cases},
    \end{equation}
    where, the implicated velocity of the real welding torch $\mathbf{V}_{t}$ due to the velocity of the robot end effector $\mathbf{V}_{r}$ is projected along the normal direction and tangential direction at the point of the welding torch. $\Vert \mathbf{d} \Vert$ represents the distance between the robot end effector and welding torch in the defined 2D plane, while $\omega$ denotes the angular velocity of the robot end effector.

    The requirement for the welding torch velocity to align with the refined welding path dictated by the SRM tip implies that the normal component of the torch's implicated velocity should be zero. This relationship can be expressed as:
    \begin{equation}\label{eq:velocity_constraint}
        \mathbf{V}_{t}^{n_{d}}\cdot{\Vec{\hat{n}}_{t}} + \mathbf{V}_{t}^{t_{d}}\cdot{\Vec{\hat{n}}_{t}} = 0,
    \end{equation}
    where, $\Vec{\hat{n}}_{t}$ denotes the normal direction at the refined path point of the welding torch. The algorithm of online robotic welding trajectory generation enhanced by tactual weld seam tracking is shown in Algorithm \ref{alg:Online_trajectory_generation}.
    \begin{algorithm}[htbp]
        \caption{Online welding trajectory generation via multi-modal fusion}
        \label{alg:Online_trajectory_generation}   
        \begin{algorithmic}[1]
            \Require {Welding seam path points $\{\mathbf{P}^{1}_{v},\mathbf{P}^{2}_{v},...,\mathbf{P}^{n}_{v}\}$ planned from 3D vision sensor, servo parameter $k_p$;}

            \State {Initialize desired deformation along normal direction $\bm{\delta}_{d}$;}
       
            \While { True }
                \State{
                $\alpha\leftarrow$ current rotation angle of the robot end effector;
                }
                
                \State{
                $\Vec{n}_{d} , \Vec{t}_{d} \leftarrow$ orthonormal vectors of touch-based welding tool;
                }
                
                \State {
                $\mathbf{P}_{r} \leftarrow$ current position of robot end effector; 
                }

                \State{
                    $\bm{\delta}^{n_s},\bm{\delta}^{t_s} \leftarrow$ current Aruco marker displacement along each direction;
                }
                \State{
                    $\{\mathbf{\hat{P}}_{t}^{1},\mathbf{\hat{P}}_{t}^{2},...,\mathbf{\hat{P}}_{t}^{m}\} \leftarrow$ \MakeUppercase{refine}($\mathbf{P}_{r},\bm{\delta}^{n_s},\bm{\delta}^{t_s}$);
                }\Comment{refined path points}

                \State{
                    $\mathbf{V}_{r}\leftarrow$\MakeUppercase{servo}($\mathbf{P}_{r},\bm{\delta}^{n_s}$);
                }\Comment{linear velocity of robot end effector}
                
                \State {
                $\mathbf{P}_{t} \leftarrow \mathbf{P}_{r} - \Vert \mathbf{d} \Vert  \Vec{t}_{d}$; 
                }\Comment{position of torch}
                
                \State{
                $\mathbf{\hat{P}}_{t} \leftarrow \mathop{\arg\min}\limits_{j\in{ \{1,2,...,m\} }}\Vert \mathbf{P}_{t} - \mathbf{\hat{P}}_{t}^{j} \Vert $
                }\Comment{nearest point to $\mathbf{P}_{t}$ in refined path}
       
                \State{
                $\Vec{\hat{n}}_{t},\Vec{\hat{t}}_{t} \leftarrow$ orthonormal vectors at point $\mathbf{\hat{P}}_{t}$
                }

                \State{
                $\mathbf{V}^{n_{d}}_{r},\mathbf{V}^{t_{d}}_{r}\leftarrow$ project $\mathbf{V}_{r}$ along $\Vec{n}_{d},\Vec{t}_{d}$
                }

                \State{
                $\omega \leftarrow -\frac{\mathbf{V}^{t_{d}}_{r}\cdot{\Vec{\hat{n}}_{t}} +  \mathbf{V}^{n_{d}}_{r}\cdot{\Vec{\hat{n}}_{t}} }{\Vert \mathbf{d} \Vert\,\Vec{n}_{d}\cdot{\Vec{\hat{n}}_{t}}}$  
                }\Comment{angular velocity of robot end effector}

                \State{
                apply velocity $(\mathbf{V}_r,\omega)$ to robot end effector;  
                }
            
             \EndWhile
            
            \Function{refine}{$\mathbf{P}_{r},\bm{\delta}^{n_s},\bm{\delta}^{t_s}$}
 
                            \State{
                            $\mathbf{P}_{s} \leftarrow \mathbf{P}_{r}^{n_s}-\bm{\delta}^{n_s}+\mathbf{P}_{r}^{t_s}-\bm{\delta}^{t_s}$ 
                            }\Comment{refine welding position vector}

                            \State{
                            $\mathbf{\hat{P}}_{t}^{j} \leftarrow \mathbf{P}_{s}$
                            }\Comment{append point to refined path 
                            
                            \State{
                            \Return $\{\mathbf{\hat{P}}_{t}^{1},\mathbf{\hat{P}}_{t}^{2},...,\mathbf{\hat{P}}_{t}^{j}\}$}
                            }
            \EndFunction
                
            \Function{servo}{$\mathbf{P}_{r},\bm{\delta}^{n_s}$}
                        \State{
                            $\mathbf{P}_{v} \leftarrow  \mathop{\arg\min}\limits_{i\in{ \{1,2,...,n\} }}\Vert \mathbf{P}_{r} - \mathbf{P}^{i}_{v} \Vert $ 
                            }\Comment{nearest point to $\mathbf{P}_{r}$ in planned path}

                        \State{
                            $\Vec{n}_{v} , \Vec{t}_{v} \leftarrow$ orthonormal vectors at point $\mathbf{P}_{v}$
                        }

                        \State{
                            $\mathbf{V}_{r} \leftarrow V_{const}\,\Vec{t}_{v} -k_{p}(\bm{\delta}^{n_s} - \bm{\delta}_{d})$
                        }\Comment{resultant linear velocity}

                        \State{
                            \Return $\mathbf{V}_{r}$
                            }

            \EndFunction
            
        \end{algorithmic}
    \end{algorithm}

\section{Results}
\label{sec:Results}

\subsection{Simulation Results for Flange-based Hand-Eye Calibration}

    Our simulation of the Eye-on-Base configuration in the Gazebo involves a 6-DOF robot of UR5 and a depth camera. The ground-true value of the hand-eye transformation matrix is set as the following, similar to a real scenario, including a roll-pitch-yaw of $[3.1415,0,-1.57]$ in radians between the robot base and camera and a translational vector of $[0.6,-0.0125,1]$ in meters.
    \begin{equation}
        _{\text{Cam}}^{\text{Base}}H_{\text{true}} = \left[\begin{array}{cccc}
        0 & -1 & 0 & 0.6\\
        -1 & 0 & 0 & -0.0125\\
        0 & 0 & -1 & 1\\
        0 & 0 & 0 & 1
        \end{array}\right].
        \label{eq:GroundTrue}
    \end{equation}

    A total sample of seventy-five tool flange poses was recorded to test the robustness of the proposed method. The robot's tool flange moved in a grid pattern within a workspace of $0.3$ m $\times$ $0.3$ m $\times$ $0.2$ m with random orientations. The actual positions of the TCP of the robot concerning the camera were obtained using TCP concerning the robot base and the ground true hand-eye transformation matrix. To investigate the proposed iterative calibration method under various levels of disturbance to the 3D scanner, different Gaussian noise $N(0,\sigma_{\text{noise}})$ were added directly to the actual values. The disturbed point is defined by
    \begin{equation}
        ^{\text{Cam}}p_{i} = {}_{\text{Base}}^{\text{Cam}}R_{\text{true}}\cdot{}^{\text{Base}}p_{i} + {}_{\text{Base}}^{\text{Cam}}t_{\text{true}} + N(0,\sigma_{\text{noise}}),
        \label{eq:noise}
    \end{equation}
    where $_{\text{Base}}^{\text{Cam}}R_{\text{true}}$ and $_{\text{Base}}^{\text{Cam}}t_{\text{true}}$ are the rotational and translational parts of the ground true hand-eye transformation matrix. The standard deviation of the Gaussian noise $\sigma_{\text{noise}}$ ranged from 0.2 mm to 10 mm at a step of 0.2 mm, representing the range of precision of industrial-grade and consumer-grade 3D scanners. For each level of Gaussian noise, the calibration results were evaluated over 100 random realizations of the noise. The rotation error $\delta R$ is expressed in terms of roll, pitch, and yaw vectors $(\delta_{\text{roll}},\delta_{\text{pitch}},\delta_{\text{yaw}})$. The ICP error metric can be directly calculated by $e_{\text{icp}} = {}_{\text{Base}}^{\text{Cam}'}\hat{H}\cdot{}_{\text{Cam}}^{\text{Base}}H_{\text{true}}$ and the cost metric is defined as the Euclidean norm of the translation error vector $\left\Vert e_{\text{icp}}\right\Vert _{\text{cost}} = \left\Vert \delta t\right\Vert _{2} = \left\Vert (\delta_{x},\delta_{y},\delta_{z})\right\Vert _{2}$.

    \begin{figure}[b!]
        \begin{centering}
        \includegraphics[width=0.9\linewidth]{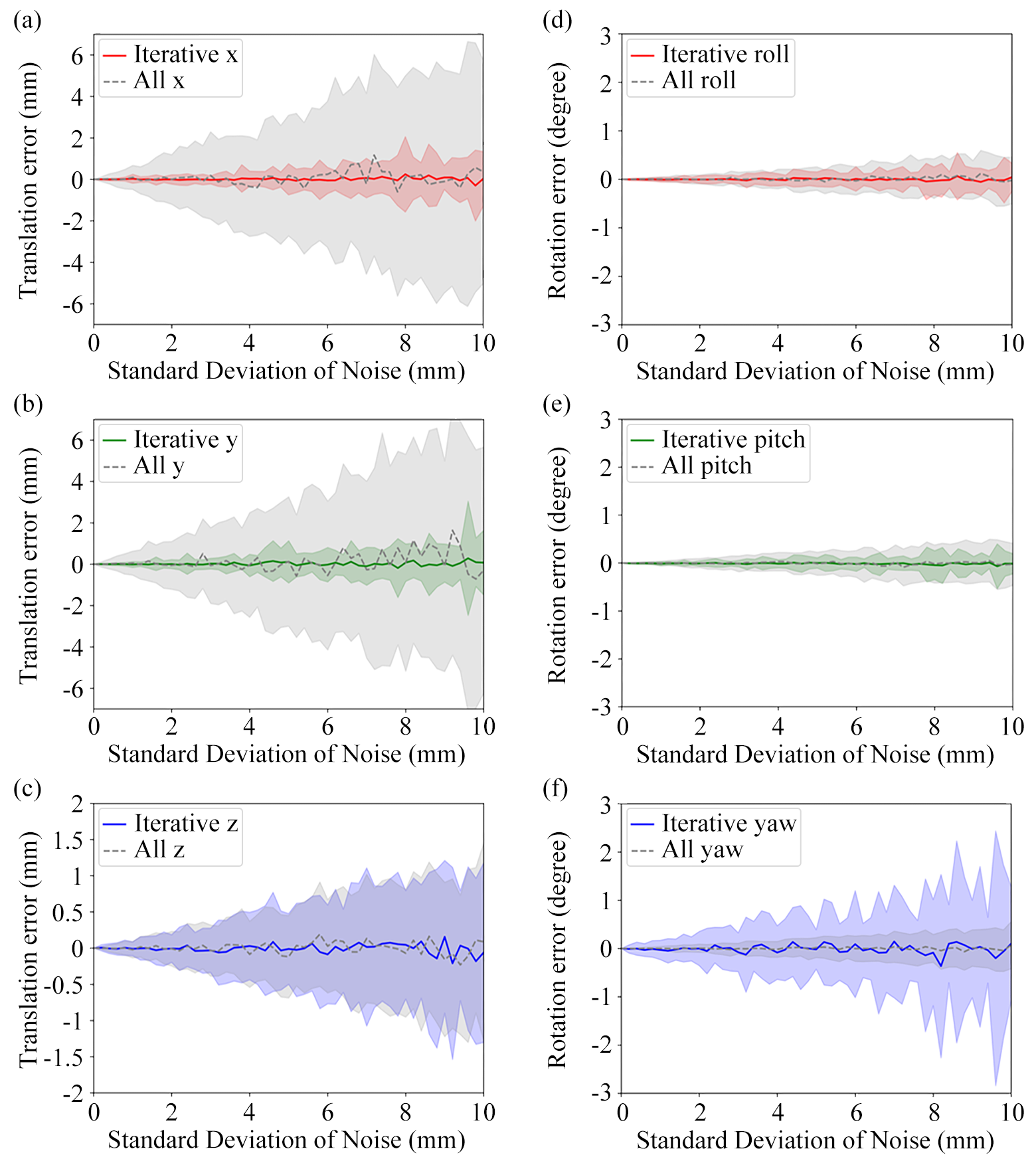}
        \par\end{centering}
        \caption{Comparison of the calibration error of using all data points (gray) and that of the proposed iterative method (colored).} 
        \label{fig:SimulationCompare}
    \end{figure}

    The results in Fig. \ref{fig:SimulationCompare} show the calibration error results using all data points versus the proposed iterative method, in which the translational and rotational errors are plotted separately concerning the various noise levels. The gray dashed line and gray shaded area denote the mean and standard deviation of the calibration error metric using all 75 data points evaluated over 100 random realizations of Gaussian noises. The solid colored line and shaded area denote the mean and standard deviation of the calibration error metric using the proposed iterative method. All mean values of the calibration errors remain close to zero as there is no systematic error. The standard deviations of the calibration errors for both methods grow linearly concerning the amplitude of the Gaussian noise. However, as shown in Figs. \ref{fig:SimulationCompare}(a-c), the iterative method is much less sensitive to noise, and the growth rate of the standard deviation of the translational error is reduced to 1/7 compared to using all data. It was also found that the translational errors are mainly contributed to by the $x$ and $y$ components. Hence, when minimizing the Euclidean norm of the translational error, the $z$-axis translational errors of both methods remain at the same level. Figs. \ref{fig:SimulationCompare}(d-f) show the yaw component mainly contributes to that rotational error. 
    
\begin{figure}[htbp]
        \begin{centering}
        \includegraphics[width=0.9\linewidth]{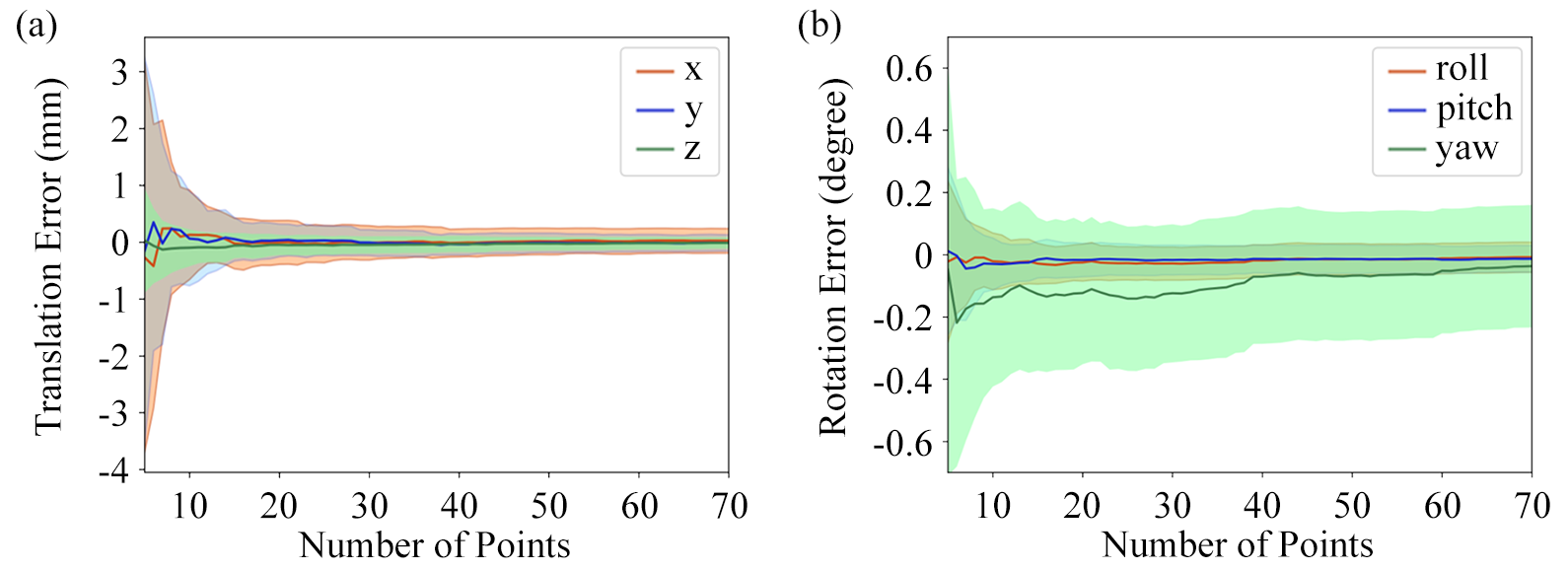}
        \par\end{centering}
        \caption{The calibration error metric concerning the iterative steps.} 
        \label{fig:SimulationError}
    \end{figure}

    Fig. \ref{fig:SimulationError} shows the calibration errors concerning the iterative steps when $\sigma_{\text{noise}}$ is set at 1 mm, and all the means of the calibration errors remain close to zero. The results are evaluated over 100 random realizations. and the solid colored line and shaded area denote the mean and standard deviation of the calibration errors.
    
    In Fig. \ref{fig:SimulationError}a, all standard deviations of the translation errors converge to less than 1 mm within ten sampled points and finally converge to 0.2 mm within 20 sampled points. In Fig. \ref{fig:SimulationError}b, all the standard deviations of the rotational errors converge to less than half of the initial values even though only the translational errors are optimized. This supports that translational and rotational errors are coupled, meaning that the decrease of one will lead to the decline of the other. The primary rotational error is a yaw angle of about 0.2 degrees. This corresponds to a 1 mm displacement in the $x$-$y$ plane at a distance $r$ of 300 mm away from the optical axis in our simulation setup, which is estimated as $\delta_{\text{yaw}}\cdot r$.

\subsection{Hardware Results for Flange-based Hand-Eye Calibration}

\subsubsection{Setup}

    The proposed iterative calibration method was applied to the hand-eye calibration between two industrial-grade 3D scanners (PhoXi S model and PhoXi M model) and four robot arms consisting of UR5, UR10e, AUBO i5, and Franka Emika. The Phoxi series scanners are based on structured light and produce up to 3.2 million 3D points. The working distances of PhoXi S and M are 384 to 520 mm and 458 to 1,118 mm, respectively. Their temporal noise and calibration accuracy are 0.05 mm and 0.1 mm, respectively. The 3D scanner was mounted about 1 meter above the table facing downward, and the robot arms were mounted on the table. The 3D scanner and robot arms were controlled by a laptop, which was also responsible for collecting and processing all the data. The computer has an Intel 2.80 GHz i7 CPU and 16 GB RAM.

    During the experiment, the robot arms moved in a grid pattern in random orientations within the workspace while the roll, pitch, and yaw concerning the robot base were all less than 0.3 radians. After segmentation and RANSAC fitting, the TCP positions in the 3D scanner frame were computed from the point cloud. We collected 60 to 80 point clouds for each robot arm, and the valid pairs of point cloud and robot pose are slightly less than these numbers as the flange segmentation failed occasionally. For analysis, we collected all the data and calculated the hand-eye calibration. However, in practice, the user should use the iterative calibration method online and do the calibration while collecting data. We shuffled the order of point clouds 50 times for each robot arm to mimic different data collection processes in which ill-conditioned points came at different sequences. The calibration results were then evaluated over these 50 calculations.

\subsubsection{Raw Data Processing}

    During the segmentation process for UR5, three false segmentations of the tool flange were purposely retained in the calibration point cloud set to demonstrate the influence of significant outliers on the performance of the proposed iterative calibration methods. The false segmented point clouds correspond to the other parts of the robot arm instead of the tool flange.

    The quality of the point cloud of tool flanges varied among robot arms. The point clouds of UR's tool flange were of good quality. For AUBO-i5, part of the tool flange was missing in some point clouds, partly due to its black color finish. Franka Emika's point clouds were incomplete due to the unique interface design. However, the RANSAC circle fitting algorithm was robust even for incomplete point clouds of tool flanges. The parameters of the RANSAC are as follows: The sample size was three, which is the minimum number to define a circle. The radius of flange $R=31$ mm. Distance threshold $e_{d}=0.3$ mm, which should be appropriately chosen according to the precision of the 3D scanner. Radius error tolerance $e_{r}=1$ mm, max iteration $k_{\text{max}}=10,000$.

\subsubsection{Results}

    Table \ref{tab:ExperimentResults} lists the calibration results with all the sample points using the SVD method. In the UR5 case, where the calibration error is so significant that the ICP algorithm fails, the translation between the TCP position measured from the point cloud and the TCP read from the robot arm controller was calculated first. This translation was used as the initial guess for the transformation of the ICP algorithm. Although only 3 out of 54 sample point clouds are false segmentations, the translation error is over 10 mm, leading to a practically unusable hand-eye matrix. Franka Emika's calibration errors are also significant, although all the flanges are correctly segmented. This is due to the geometry feature of Franka Emika's flange, where the circle fitting is more challenging. The calibration results of UR10e are relatively better, with translational errors of less than 2 mm. AUBO i5 and Phoxi S achieved the best calibration results with less than 0.4 mm translation errors. The tool flange of Franka Emika and that of AUBO i5 have the same geometry and are correctly segmented in the experiment. The calibration results from AUBO i5 and Phoxi S suggest that 3D scanners with higher precision lead to higher calibration accuracy.
    \begin{table}[htbp]
        \centering
        \caption{Calibration errors using all sample points without iterative optimization.}
        \label{tab:ExperimentResults}
        \begin{tabular}{ccccccc}
            \hline
            \begin{tabular}[c]{@{}c@{}}Robot Setups\end{tabular} &
            \begin{tabular}[c]{@{}c@{}}$x$\\ (mm)\end{tabular} &
            \begin{tabular}[c]{@{}c@{}}$y$\\ (mm)\end{tabular} &
            \begin{tabular}[c]{@{}c@{}}$z$\\ (mm)\end{tabular} &
            \begin{tabular}[c]{@{}c@{}}$\text{roll}$\\ ($^{\circ}$)\end{tabular} &
            \begin{tabular}[c]{@{}c@{}}$\text{pitch}$\\ ($^{\circ}$)\end{tabular} &
            \begin{tabular}[c]{@{}c@{}}$\text{yaw}$\\ ($^{\circ}$)\end{tabular} \\ \hline
            \begin{tabular}[c]{@{}c@{}}UR5 \& PhoXi M\end{tabular}     & 11.60 & 2.81  & 5.55  & 0.49  & -0.89 & -0.47 \\
            \begin{tabular}[c]{@{}c@{}}UR10e \& PhoXi M\end{tabular}   & 0.60  & 1.75  & -0.41 & 0.10  & -0.04 & -0.23 \\
            \begin{tabular}[c]{@{}c@{}}Franka Emika \& PhoXi M\end{tabular}  & 4.34  & -4.06 & -0.13 & -0.30 & -0.34 & -0.07 \\
            \begin{tabular}[c]{@{}c@{}}AUBO i5 \& PhoXi S\end{tabular} & -0.26 & -0.27 & -0.35 & 0.14  & 0.04  & 0.01  \\ \hline
        \end{tabular}
    \end{table}
    
\begin{figure}[htbp]
        \begin{centering}
        \includegraphics[width=0.9\linewidth]{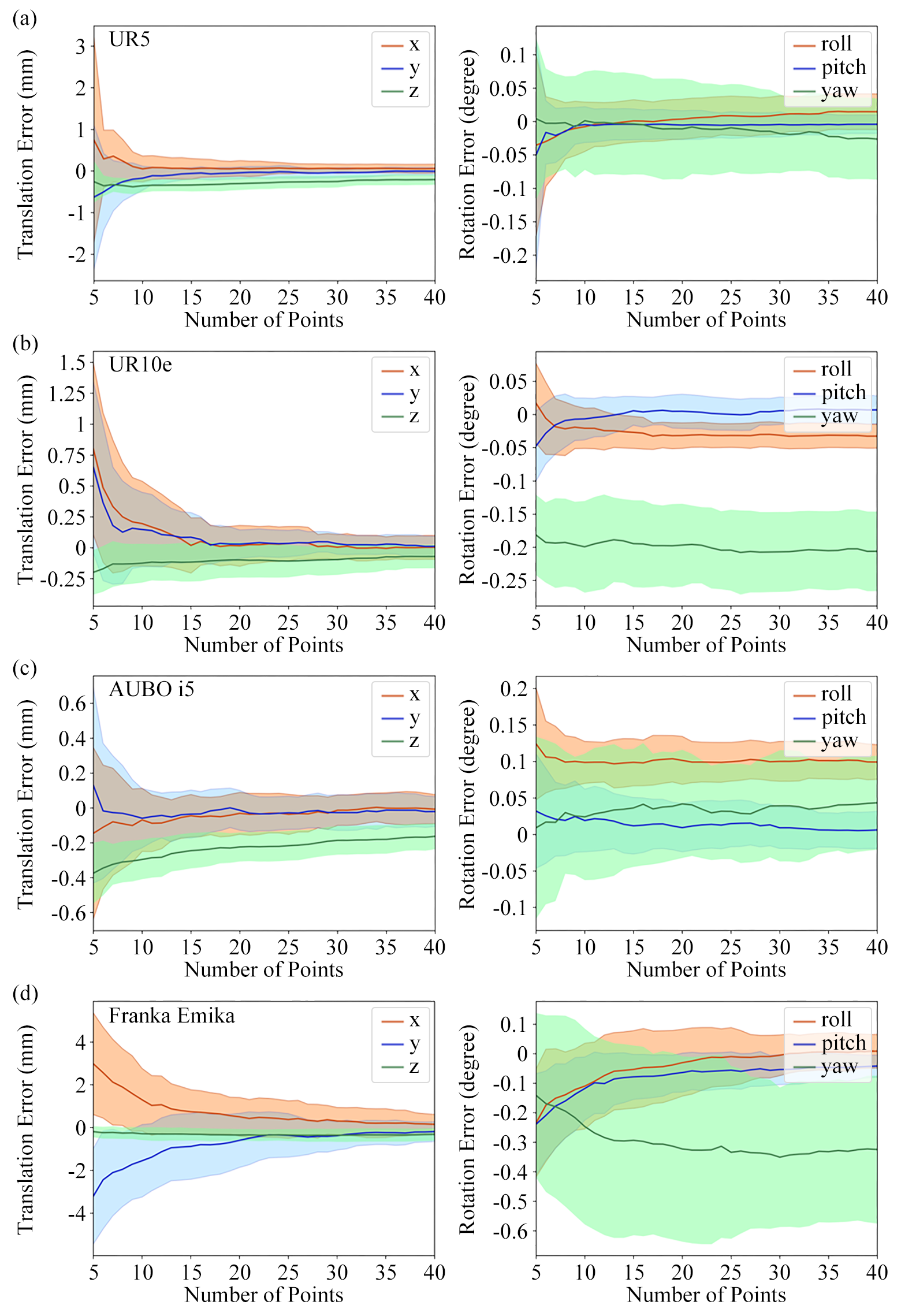}
        \par\end{centering}
        \caption{The calibration error metric concerning the iterative steps.} 
        \label{fig:experimentalErrors}
    \end{figure}

    Fig. \ref{fig:experimentalErrors} shows the calibration errors concerning the iterative steps for robot UR5, UR10e, AUBO i5, and Franka Emika. The results are evaluated over 50 calculations, and means are plotted in lines and standard deviations as shaded areas. In all four scenarios, the translation errors are mainly contributed by $x$ and $y$ components, and the rotation errors are primarily by the yaw component, which agrees well with the simulation results. UR5, UR10e, and AUBO i5 converge to less than 0.28 mm, while their rotational errors converge to less than 0.25 degrees. Despite the challenge in circle fitting, the translation errors of Franka Emika converge to less than 0.4 mm at the expense of growth in yaw-component rotation error converging to less than 0.6 degrees.

\subsubsection{Accuracy and Efficiency}

    The hand-eye calibration accuracy based on 3D geometry is highly dependent on the precision of 3D scanners, the geometry of the tool flanges, and the robustness of the feature extraction process. The proposed iterative calibration method achieves much better accuracy than only using point data, especially when there are significant outliers in the data. The convergence speed and amplitude of calibration errors depend on the point cloud quality of the tool flange and the robustness of the circle fitting algorithms. The iterative method is statistically meaningful as it resists various noise levels. In practice, a more efficient online iterative calibration process is to perform error compensation on the calculated hand-eye matrix once a valid ICP error metric is obtained in steps 3 or steps 12 and 13 in Algorithm \ref{alg:IterativeCalibration}. The error compensation is formulated as
    \begin{equation}
        \hat{H}_{\text{compensated}} = \hat{H}_{\text{optimal}}\cdot e_{\text{optimal}}.
        \label{eq:Hcompensated}
    \end{equation}
    
    In the experiment results, the initial $e_{\text{optimal}}$ in step 3 in Algorithm \ref{alg:IterativeCalibration} is sometimes infinite due to noisy data. However, after the first iteration, the chance of $e_{\text{optimal}}$ remaining infinity is found to be small. In Fig. \ref{fig:experimentalErrors}, all 50 random calibration calculations conducted for each robot arm have finite $e_{\text{optimal}}$ within five pairs of data, which can be referred from the limited means and standard deviations of components of $e_{\text{optimal}}$.
    
    We also included the hand-eye calibration result between the 3D scanner PhoXi M and UR10e using commercial software developed by Photoneo. The software uses a calibration sphere mounted on the robot tool flange and calculates the hand-eye matrix after collecting a minimum of four pairs of robot pose and point cloud. A few calibration trials with 4 and 16 pairs of data are conducted respectively for comparison, and their calibration errors are calculated using the ICP metric defined in Eq. \eqref{eq:ErrorMetric} with the same verification point cloud $^{\text{Cam}}P_{v}$. The results in Table \ref{tab:CalibErrorsPhotoneo} show that more sample points lead to fewer errors. 
    \begin{table}[htbp]
        \centering
        \caption{The calibration errors using Photoneo software with a spherical calibration object.}
        \label{tab:CalibErrorsPhotoneo}
        \begin{tabular}{ccccccc}
            \hline
            \begin{tabular}[c]{@{}c@{}}Number \\ of points\end{tabular} &
              \begin{tabular}[c]{@{}c@{}}$x$\\ (mm)\end{tabular} &
              \begin{tabular}[c]{@{}c@{}}$y$\\ (mm)\end{tabular} &
              \begin{tabular}[c]{@{}c@{}}$z$\\ (mm)\end{tabular} &
              \begin{tabular}[c]{@{}c@{}}$\text{roll}$\\ ($^{\circ}$)\end{tabular} &
              \begin{tabular}[c]{@{}c@{}}$\text{pitch}$\\ ($^{\circ}$)\end{tabular} &
              \begin{tabular}[c]{@{}c@{}}$\text{yaw}$\\ ($^{\circ}$)\end{tabular} \\ \hline
            4 pairs &
              -1.94 &
              -0.90 &
              -0.86 &
              -0.07 &
              0.04 &
              -0.10 \\
            16 pairs &
              -0.98 &
              -1.01 &
              -0.83 &
              -0.07 &
              0.09 &
              -0.11 \\ \hline
            \end{tabular}
    \end{table}
    However, it is also found that the calibration result is sensitive to the diversity of the sampled points. In one of the few trials where the sample points did not vary enough, the translational error is about 6 mm, which is unusable.

\subsection{Demonstrated Applications in Soft Robotic Tactile Welding}
\label{sec:Welding}

\begin{figure}[htbp]
        \centering
        \includegraphics[width=0.7\linewidth]{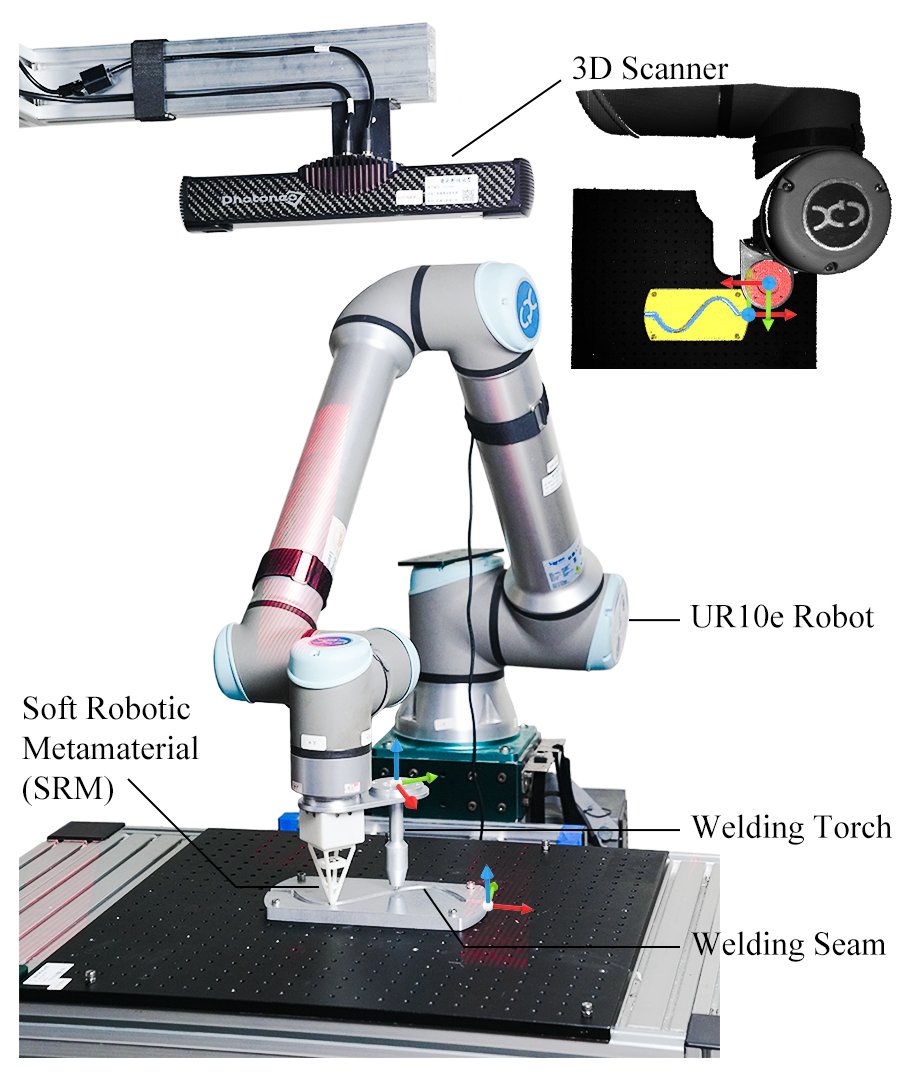}
        \caption{Experiment setup of flange-based calibration for tactile-enhanced robotic welding.} 
        \label{fig:Welding}
    \end{figure}

    This section presents a robotic welding system within a laboratory environment, as shown in Fig. \ref{fig:Welding}. The hardware platform comprises a PhoXi M 3D scanner (resolution at 0.3 mm), a collaborative robot (UR10e, repeatability accuracy at 0.05 mm), and the proposed touch-based welding tool mounted to the robot's end-effector. All hardware is connected to a computer to process point clouds, real-time tactile signals, and robot control.
    
    In Fig. \ref{fig:Welding}, the system leverages the proposed flange-based method to obtain the hand-eye transformation. After capturing the workpiece's point cloud from the 3D scanner, we registered it with the workpiece's CAD model to get the position of the welding seam in the scanner's coordinate. The position is then converted to the robot base's coordinate using the hand-eye transformation. The SRM tip is designed to glide ahead of the welding torch, tracing the weld seam extracted and planned by the 3D scanner. Utilizing the deformation servo mechanism featured in Section \ref{sec:welding_seam_tracking}, the system corrects the path followed by the robot end effector, applying an offset $\bm{\delta}$ that compensates for the deformation detected by the SRM at the relevant location. Consequently, the welding torch is instructed to carry out the adjusted trajectory. 
    
\begin{figure}[htbp]
        \centering
        \includegraphics[width=0.8\linewidth]{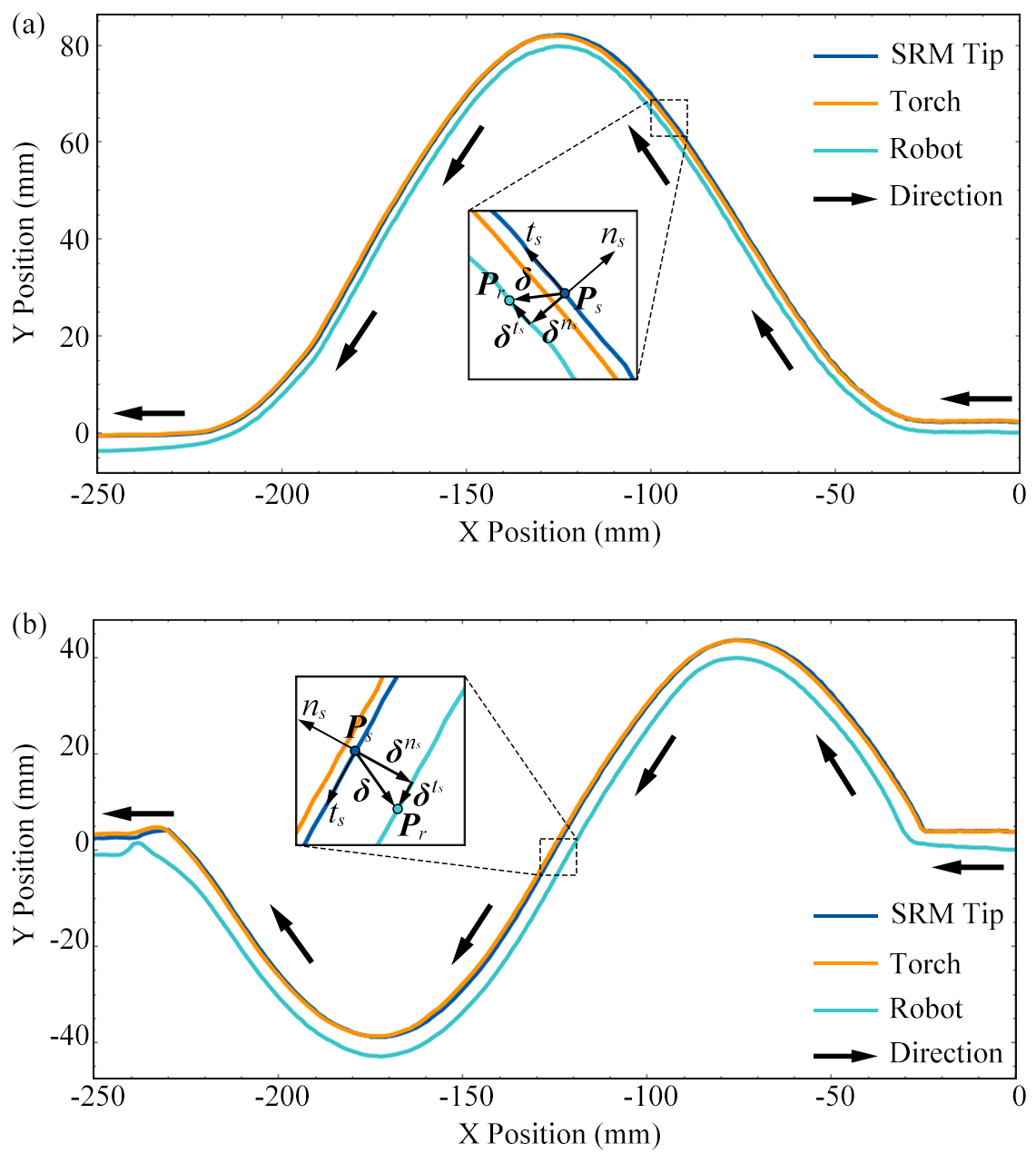}
        \caption{Executed paths of SRM tip, welding torch, and robot for (a) and (b)two different welding paths.}
        \label{fig:WeldingPath}
    \end{figure}

    Fig. \ref{fig:WeldingPath} shows the SRM tip, welding torch, and robot paths. During the welding experiment, corrective adjustments are limited to those within a two-dimensional plane orthogonal to the z-axis. The corrective offset in the direction perpendicular to the SRM tip $\bm{\delta}^{n_s}$ is maintained at a nearly constant magnitude throughout the trajectory, in alignment with the initially specified value $\bm{\delta}_{d}$. An exception occurs at the end of the welding path, where the SRM unexpectedly loses contact with the seam, prompting us to reset the deformation servo reference value $\bm{\delta}_{d}$ to zero. This adjustment results in a noticeable jitter at the end portion of the path, as seen in the plot.

    Assisted by the straightforward flange hand-eye calibration method and tactile information provided by the low-cost SRM, the robotic welding system accomplishes autonomous weld seam tracking for two distinct workpieces. See Movie S1 in the Supplementary Materials for a video demonstration. 

\section{Discussion}
\label{sec:Discussion}

\subsection{Towards a Depth-based Hand-Eye Calibration}

\begin{figure}[htbp]
        \centering
        \includegraphics[width=\linewidth]{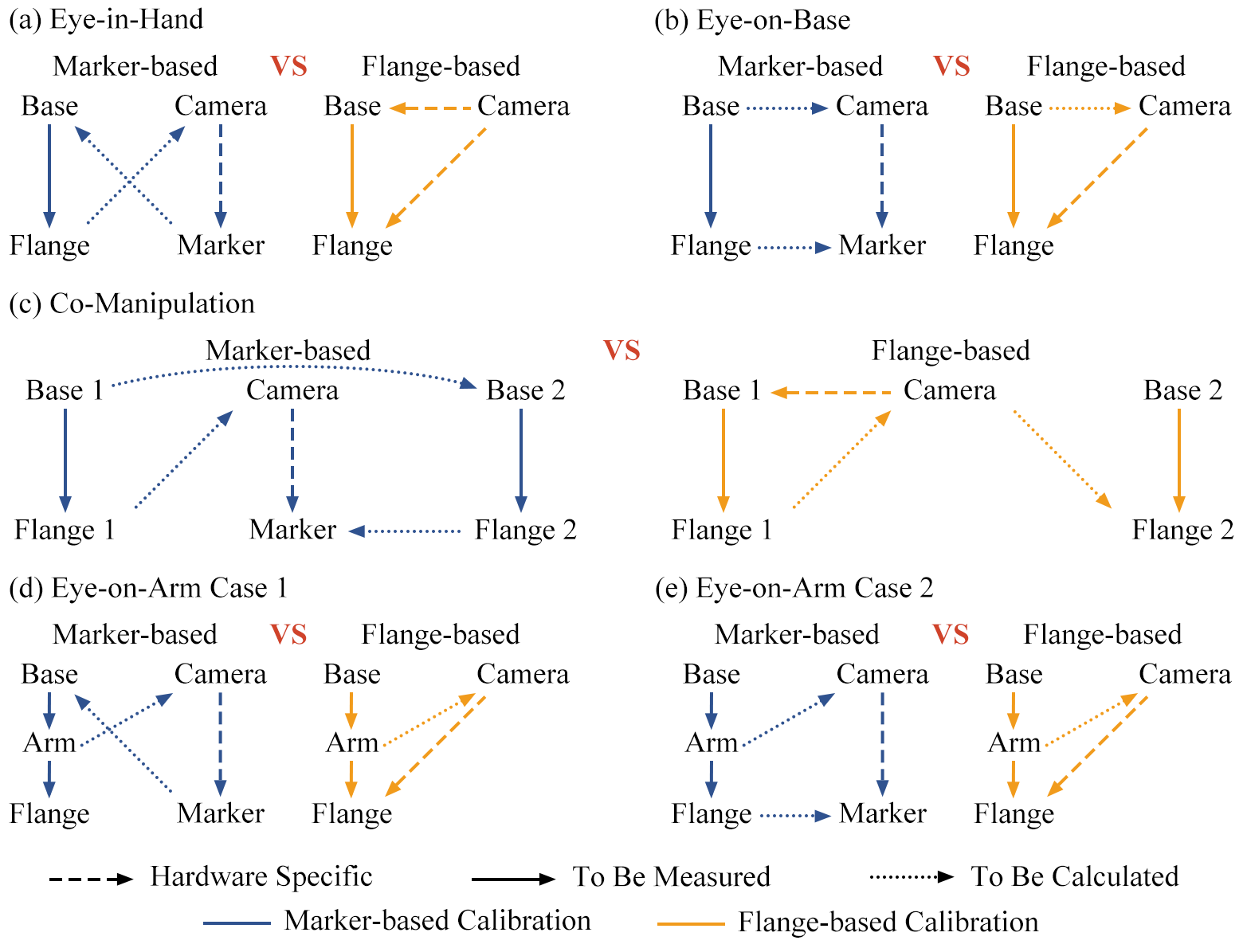}
        \caption{A graphical summary of hand-eye calibration methods with common marker-based configurations and the proposed flange-based methods with updated configurations.}
        \label{fig:CalibrationReview}
    \end{figure}

    Figs. \ref{fig:CalibrationReview} shows graphical representations summarizing the above configurations of hand-eye calibration. The proposed method in this paper is to exploit high-quality, three-dimensional perception against standardized design and manufacturing of the mechanical components on the robot, which fall within a comparable level of accuracy and tolerance. Therefore, one can reconfigure these established methods for hand-eye calibration by removing the $\{ \text{Mark}\} $ frame in each case. Note that in the reconfigured Eye-in-Hand case in Fig. \ref{fig:CalibrationReview}a, the camera can alternatively inspect the geometric features, i.e., circles, on the base mounting flange commonly found in most robots to proceed with the proposed calibration method. We suggest that one can freely choose any geometric features on the robot as long as a direct reading or deduction of the referencing feature's pose information can be obtained from the robot's controller or teach pendant. 

\subsection{Robot Flanges for Hand-Eye Calibration}

     In both simulation and experiment results, the cost metric minimized was chosen as the Euclidean norm of the translational error vector. The reason is that the impact of the translational error on applying the hand-eye transformation matrix is uniform in a 3D scanner. However, the rotational error is found to be small enough, and its impact on the application is minimized around the optical axis of the 3D scanner, which linearly increases when moving away from the optical axis. In most scenarios, the workspace is usually located around the optical axis of the 3D scanner. Users can choose the appropriate elements to minimize according to their needs. For example, in the pick and place scenario, one might only minimize the translation errors in the $x$ and $y$-axis if the end-effector has flexibility in the $z$-axis. If both translation and rotation errors are critical, the $\Vert\cdot\Vert_{\text{cost}}$ can be defined as $\left\Vert (\delta_{x},\delta_{y},\delta_{z},\delta_{\text{roll}}\cdot r,\delta_{\text{pitch}}\cdot r,\delta_{\text{yaw}}\cdot r)\right\Vert _{2}$, where $r$ is the work radius from the optical axis of the 3D scanner.
     
     We also experimented with our proposed method using a commercial-grade 3D scanner to test its performance. The latest release of Microsoft's Kinect sensor, the Azure Kinect DK, contains a 1-MP time-of-flight (ToF) depth sensor. Following a similar process, we experimented using a UR10e and an Azure Kinect DK. Due to the technical limitation of the ToF mechanism on metal surfaces, the quality of the point cloud of the tool flange is rather noisy. Instead of looking for the outer circle of the tool flange, we switch to the outer circle on UR10e's wrist joint, which is 90 mm, as in Fig. \ref{fig:CobotFlanges}b. Fifty pairs of point clouds and the corresponding robot poses are recorded. Verification point clouds and robot pose are used to calculate the ICP error metric. The translational and rotational errors using all the data points are $[-11.11,3.90,7.15]$ in mm and $[-0.15,0.92,1.56]$ in degrees. The compensated hand-eye matrix $\hat{H}_{\text{compensated}}$ can be obtained from Eq. \eqref{eq:Hcompensated}. Nevertheless, with high-quality depth sensors entering consumer-grade applications, our method holds the potential for a broader range of applications, where a raw measurement of the depth data can be used directly for calibration and interaction.

\subsection{Integration of Vision and Tactile for Robotic Welding}

    In robot welding applications, hand-eye calibration is crucial to ensure precision in positioning the welding apparatus, which directly influences the quality and integrity of the welds. The integration of tactile sensors allows the welding process to receive real-time feedback, enabling the system to adjust, adapt, and refine its operations based on the feedback received. By combining global visual planning with local tactile feedback, our system achieves comprehensive supervision and control over the welding process, significantly enhancing precision and adaptability.

    The implications of our research for robotic vision and welding are multifaceted. Firstly, our findings suggest that integrating tactile feedback into visual planning systems can substantially improve weld quality, as the system can dynamically respond to variations in the welding environment. Secondly, applying machine learning algorithms to process the feedback data can further enhance the accuracy of corrective actions, leading to more consistent and reliable welds. This approach improves the current state of robotic welding systems and opens new avenues for research in adaptive control and intelligent decision-making in robotics.

    Our presented robotic welding system offers several opportunities for improvement, such as refining spatial welding paths based on the designed tactile sensor \cite{Lei2020ATactual} and improving the accuracy of corrective actions through the use of machine learning algorithms \cite{Mahadevan2021Intelligent}. These enhancements are expected to significantly advance robotic vision and welding technologies, ultimately leading to more efficient and high-quality welding processes.

\section{Conclusion, Limitations, and Future Work}
\label{sec:Conclusion}

\subsection{Conclusion}

    In this paper, we proposed an iterative hand-eye calibration method based on the 3D measurement of the robot tool flange. Using the Tool Center Point (TCP) of the robot arm as the referencing point, the hand-eye calibration is simplified to fitting two 3D point sets with the least-squares analytical solution. The proposed method adopts the 3D error metric based on Iterative Closest Point registration to monitor and optimize the online calibration process. Once the desired hand-eye calibration accuracy is achieved, the calibration process is stopped, requiring only a minimum set of point clouds to be processed. The proposed method was applied to calculate the hand-eye transformation between 3D scanners, including Photoneo Phoxi S \& M and Microsoft Azure Kinect DK, and four robot arms, including UR5, UR10e, AUBO i5, and Franka Emika. The results showed that the iterative method converged quickly and was robust with tested robots. In addition, combining the flange-based calibration method with soft tactile sensing, a welding seam tracking system was presented. The demonstration indicated that the system could accurately compute the transformation among the robot tool flange, the welding seam, and the world frame, ensuring the welding torch's movement along the correct trajectory. The integration of the soft tactile sensor enabled the robot to adjust and refine the operations following the real-time feedback. The welding system achieved more consistent welding quality through flange-based calibration and soft tactile sensing feedback. 

\subsection{Limitations}

    The proposed online welding trajectory generation Algorithm \ref{alg:Online_trajectory_generation} heavily relies on the geometric relationship between the weld seam and the welding tool. In practice, using this novel touch-based welding tool can lead to kinematic singularities when $\Vec{n}_{d}\perp{\Vec{\hat{n}}_{t}}$, making it impossible to adjust the angular velocity $\omega$ of robot end effector to meet the velocity requirements of the seam in Eq. \eqref{eq:velocity_constraint}. Furthermore, there is a concern about the sudden change in the deformation servo reference value $\bm{\delta}_{d}$, which leads to a significant deviation from the welding seam at the end of the path. These issues seem to hinder the effectiveness of the algorithm. It might be worth considering ways to address these problems to improve the quality of soft robotic tactile welding processes.
    
\subsection{Future Work}

    Our future work will investigate further details on applying the proposed approach to scenarios where the 3D scanner is mounted on the robot arm. This flexible setup allows robots to perform hand-eye calibration while moving, making it well-suited for challenging mobile welding tasks in unstructured environments. Additionally, we will further explore research into utilizing machine learning algorithms for real-time calculation of corrected welding trajectories based on the output data from SRM in more intricate scenarios.

\section*{Acknowledgment}
\label{sec:Acknowledge}

    This work was partly supported by the Science, Technology, and Innovation Commission of Shenzhen Municipality [JSGG20220831110002004], the National Natural Science Foundation of China [62206119], Shenzhen Long-Term Support for Higher Education at SUSTech [20231115141649002], and SUSTech Virtual Teaching Lab for Machine Intelligence Design and Learning [Y01331838].

\section*{Supplementary Materials}
\label{sec:Appendix}

    \begin{itemize}
        \item \textbf{Movie S1. Evaluation of tactile-enhanced autonomous weld seam tracking.} In this movie, we showcase the proposed robotic welding system achieving autonomous weld seam tracking for two different workpieces in Section \ref{sec:Welding}. 
    \end{itemize}

\bibliographystyle{unsrt}
\bibliography{References}  

\begin{thebibliography}{10}

\bibitem{Sarbolandi2015KinectRange}
Hamed Sarbolandi, Damien Lefloch, and Andreas Kolb.
\newblock \href{https://doi.org/10.1016/j.cviu.2015.05.006}{Kinect Range
  Sensing: Structured-light versus Time-of-Flight Kinect}.
\newblock {\em Computer Vision and Image Understanding}, 139:1--20, 2015.

\bibitem{Jarabo2017RecentAdvances}
Adrian Jarabo, Belen Masia, Julio Marco, and Diego Gutierrez.
\newblock \href{https://doi.org/10.1016/j.visinf.2017.01.008}{Recent Advances
  in Transient Imaging: A Computer Graphics and Vision Perspective}.
\newblock {\em Visual Informatics}, 1(1):65--79, 2017.

\bibitem{Kolb2010ToF}
Andreas Kolb, Erhardt Barth, Reinhard Koch, and Rasmus Larsen.
\newblock
  \href{https://doi.org/10.1111/j.1467-8659.2009.01583.x}{Time-of-Flight
  Cameras in Computer Graphics}.
\newblock {\em Computer Graphics Forum}, 29(1):141--159, 2010.

\bibitem{Fan2019APrecision}
Junfeng Fan, Fengshui Jing, Lei Yang, Long Teng, and Min Tan.
\newblock \href{https://doi.org/10.1109/JSEN.2018.2876144}{A Precise Initial
  Weld Point Guiding Method of Micro-Gap Weld Based on Structured Light Vision
  Sensor}.
\newblock {\em IEEE Sensors Journal}, 19(1):322--331, 2019.

\bibitem{Lu2023Automatic}
Xueqin Lü, Chengzhi Xie, Xianghuan He, Siwei Li, Yuzhe Xu, Songjie He, Jian
  Fang, Min Zhang, and Xingwu Yang.
\newblock \href{https://doi.org/10.1109/JSEN.2022.3224931}{Automatic
  Recognition of Multiple Weld Types Based on Structured Light Vision Sensor
  Using Deep Transfer Learning}.
\newblock {\em IEEE Sensors Journal}, 23(7):7142--7152, 2023.

\bibitem{Zhang2020ErrorCorrectable}
Liang Zhang, Jian-Zhou Zhang, Xiaoyi Jiang, and Biao Liang.
\newblock \href{https://doi.org/10.1109/TIM.2020.2987492}{Error Correctable
  Hand-Eye Calibration for Stripe-Laser Vision-Guided Robotics}.
\newblock {\em IEEE Transactions on Instrumentation and Measurement},
  69(10):8314--8327, 2020.

\bibitem{Peng2021AHybrid}
Jianqing Peng, Wenfu Xu, Fengxu Wang, Yu~Han, and Bin Liang.
\newblock \href{https://doi.org/10.1109/TIM.2021.3078523}{A Hybrid Hand-Eye
  Calibration Method for Multilink Cable-Driven Hyper-Redundant Manipulators}.
\newblock {\em IEEE Transactions on Instrumentation and Measurement}, 70:1--13,
  2021.

\bibitem{Shi2012LevelsOfHuman}
Jane Shi, Glenn Jimmerson, Tom Pearson, and Roland Menassa.
\newblock \href{https://doi.org/10.1145/2393091.2393111}{Levels of Human and
  Robot Collaboration for Automotive Manufacturing}.
\newblock In {\em Proceedings of the Workshop on Performance Metrics for
  Intelligent Systems}, pages 95--100, 2012.

\bibitem{Hyatt2019ConfigurationEstimation}
Phillip Hyatt, Dustan Kraus, Vallan Sherrod, Levi Rupert, Nathan Day, and
  Marc~D Killpack.
\newblock \href{https://doi.org/10.1109/TMECH.2018.2878228}{Configuration
  Estimation for Accurate Position Control of Large-Scale Soft Robots}.
\newblock {\em IEEE/ASME Transactions on Mechatronics}, 24(1):88--99, 2018.

\bibitem{Rout2019Advances}
Amruta Rout, B.B.V.L. Deepak, and B.B. Biswal.
\newblock \href{https://doi.org/10.1016/j.rcim.2018.08.003}{Advances in Weld
  Seam Tracking Techniques for Robotic Welding: A Review}.
\newblock {\em Robotics and Computer-Integrated Manufacturing}, 56:12--37,
  2019.

\bibitem{Wu2016Simultaneous}
Liao Wu, Jiaole Wang, Lin Qi, Keyu Wu, Hongliang Ren, and Max Q-H Meng.
\newblock \href{https://doi.org/10.1109/TRO.2016.2530079}{Simultaneous
  Hand-Eye, Tool-Flange, and Robot-Robot Calibration for Comanipulation by
  Solving the AXB = YCZ Problem}.
\newblock {\em IEEE Transactions on Robotics}, 32(2):413--428, 2016.

\bibitem{Wu2020HandEye}
Jin Wu, Yuxiang Sun, Miaomiao Wang, and Ming Liu.
\newblock \href{https://doi.org/10.1109/TIM.2019.2930710}{Hand-Eye Calibration:
  4-D Procrustes Analysis Approach}.
\newblock {\em IEEE Transactions on Instrumentation and Measurement},
  69(6):2966--2981, 2020.

\bibitem{Huang2023DynamicParameter}
Yanjiang Huang, Jianhong Ke, Xianmin Zhang, and Jun Ota.
\newblock \href{https://doi.org/10.1109/TRO.2022.3211194}{Dynamic Parameter
  Identification of Serial Robots Using a Hybrid Approach}.
\newblock {\em IEEE Transactions on Robotics}, 39(2):1607--1621, 2023.

\bibitem{Jiang2017KinematicAccuracy}
Yao Jiang, Tiemin Li, Liping Wang, and Feifan Chen.
\newblock \href{https://doi.org/10.1109/TMECH.2017.2756348}{Kinematic Accuracy
  Improvement of a Novel Smart Structure-Based Parallel Kinematic Machine}.
\newblock {\em IEEE/ASME Transactions on Mechatronics}, 23(1):469--481, 2017.

\bibitem{Kluz2014TheRepeatability}
Rafa{\l} Kluz and Tomasz Trzepieci{\'n}ski.
\newblock \href{https://doi.org/10.1108/AA-07-2013-070}{The Repeatability
  Positioning Analysis of the Industrial Robot Arm}.
\newblock {\em Assembly Automation}, 34(3):285--295, 2014.

\bibitem{Zhang2021ASimultaneous}
Yuan Zhang, Zhicheng Qiu, and Xianmin Zhang.
\newblock \href{https://doi.org/10.1109/TIM.2020.3013308}{A Simultaneous
  Optimization Method of Calibration and Measurement for a Typical Hand–Eye
  Positioning System}.
\newblock {\em IEEE Transactions on Instrumentation and Measurement},
  70(5002111):1--11, 2021.

\bibitem{Kahn2014HandEye}
Svenja Kahn, Dominik Haumann, and Volker Willert.
\newblock \href{https://ieeexplore.ieee.org/document/7295121}{Hand-Eye
  Calibration with a Depth Camera: 2D or 3D?}
\newblock In {\em International Conference on Computer Vision Theory and
  Applications (VISAPP)}, volume~3, pages 481--489. IEEE, 2014.

\bibitem{Shah2012AnOverview}
Mili Shah, Roger~D Eastman, and Tsai Hong.
\newblock \href{https://doi.org/10.1145/2393091.2393095}{An Overview of
  Robot-Sensor Calibration Methods for Evaluation of Perception Systems}.
\newblock In {\em Proceedings of the Workshop on Performance Metrics for
  Intelligent Systems}, pages 15--20. ACM, 2012.

\bibitem{Wang2022Robot}
Xiao Wang, Kuanyong Zhou, Jianning Yang, and Hanwen Song.
\newblock \href{https://doi.org/10.1016/j.measurement.2022.112076}{Robot-world
  and hand–eye calibration based on motion tensor with applications in
  uncalibrated robot}.
\newblock {\em Measurement}, 204:112076, 2022.

\bibitem{Hu2013ARapid}
Fengjun Hu.
\newblock \href{https://doi.org/10.12733/JICS20101595}{A Rapid Eye-to-Hand
  Coordination Method of Industrial Robots}.
\newblock {\em Journal of Information \& Computational Science},
  10(5):1489--1496, 2013.

\bibitem{Zhang2017StereoVision}
Xuanchen Zhang, Yuntao Song, Yang Yang, and Hongtao Pan.
\newblock \href{https://doi.org/10.1016/j.robot.2017.04.001}{Stereo Vision
  based Autonomous Robot Calibration}.
\newblock {\em Robotics and Autonomous Systems}, 93:43--51, 2017.

\bibitem{Yang2018RoboticHandEye}
Lixin Yang, Qixin Cao, Minjie Lin, Haoruo Zhang, and Zhuoming Ma.
\newblock \href{https://doi.org/10.1109/ICCAR.2018.8384652}{Robotic Hand-Eye
  Calibration with Depth Camera: A Sphere Model Approach}.
\newblock In {\em International Conference on Control, Automation and Robotics
  (ICCAR)}, pages 104--110. IEEE, 2018.

\bibitem{Li2018Simultaneous}
Wei Li, Mingli Dong, Naiguang Lu, Xiaoping Lou, and Peng Sun.
\newblock \href{https://doi.org/10.3390/s18113949}{Simultaneous Robot–World
  and Hand–Eye Calibration without a Calibration Object}.
\newblock {\em Sensors}, 18(11):3949, 2018.

\bibitem{An2016MethodFor}
Yatong An, Tyler Bell, Beiwen Li, Jing Xu, and Song Zhang.
\newblock \href{https://doi.org/10.1364/AO.55.009563}{Method for Large-Range
  Structured Light System Calibration}.
\newblock {\em Applied Optics}, 55(33):9563--9572, 2016.

\bibitem{kalia2019marker}
Megha Kalia, Prateek Mathur, Nassir Navab, and Septimiu~E Salcudean.
\newblock \href{https://doi.org/10.1049/htl.2019.0094}{Marker-less real-time
  intra-operative camera and hand-eye calibration procedure for surgical
  augmented reality}.
\newblock {\em Healthcare technology letters}, 6(6):255--260, 2019.

\bibitem{mcgarry2022assessment}
Lauren McGarry, Joseph Butterfield, and Adrian Murphy.
\newblock \href{https://doi.org/10.1016/j.rcim.2021.102275}{Assessment of ISO
  Standardisation to Identify an Industrial Robot's Base Frame}.
\newblock {\em Robotics and Computer-Integrated Manufacturing}, 74:102275,
  2022.

\bibitem{ISO9409-1}
ISO.
\newblock \href{https://www.iso.org/standard/36578.html}{ISO 9409-1:2004 -
  Manipulating industrial robots -- Mechanical interfaces -- Part 1: Plates}.
\newblock Technical report, International Organization for Standardization,
  2004.

\bibitem{Geng2024ANovel}
Yusen Geng, Yuankai Zhang, Xincheng Tian, and Lelai Zhou.
\newblock \href{https://doi.org/10.1016/j.rcim.2023.102702}{A Novel 3D
  Vision-based Robotic Welding Path Extraction Method for Complex Intersection
  Curves}.
\newblock {\em Robotics and Computer-Integrated Manufacturing}, 87:102702,
  2024.

\bibitem{Geng2023ANovel}
Yusen Geng, Min Lai, Xincheng Tian, Xiaolong Xu, Yong Jiang, and Yuankai Zhang.
\newblock \href{https://doi.org/10.1016/j.rcim.2022.102433}{A Novel Seam
  Extraction and Path Planning Method for Robotic Welding of Medium-Thickness
  Plate Structural Parts based on 3D Vision}.
\newblock {\em Robotics and Computer-Integrated Manufacturing}, 79:102433,
  2023.

\bibitem{Lei2020ATactual}
Ting Lei, Yu~Huang, Wenjun Shao, Weinan Liu, and Youmin Rong.
\newblock \href{https://doi.org/10.1016/j.rcim.2019.101864}{A Tactual Weld Seam
  Tracking Method in Super Narrow Gap of Thick Plates}.
\newblock {\em Robotics and Computer-Integrated Manufacturing}, 62:101864,
  2020.

\bibitem{Michael2020HapticBased}
Michael Tannous, Marco Miraglia, Francesco Inglese, Luca Giorgini, Filippo
  Ricciardi, Riccardo Pelliccia, Mario Milazzo, and Cesare Stefanini.
\newblock \href{https://doi.org/10.1016/j.rcim.2020.101952}{Haptic-based Touch
  Detection for Collaborative Robots in Welding Applications}.
\newblock {\em Robotics and Computer-Integrated Manufacturing}, 64:101952,
  2020.

\bibitem{suwanratchatamanee2009robotic}
Kitti Suwanratchatamanee, Mitsuharu Matsumoto, and Shuji Hashimoto.
\newblock \href{https://doi.org/10.1109/TIE.2009.2031195}{Robotic tactile
  sensor system and applications}.
\newblock {\em IEEE Transactions on Industrial Electronics}, 57(3):1074--1087,
  2009.

\bibitem{lepora2021soft}
Nathan~F Lepora.
\newblock \href{https://doi.org/10.1109/JSEN.2021.3100645}{Soft biomimetic
  optical tactile sensing with the TacTip: A review}.
\newblock {\em IEEE Sensors Journal}, 21(19):21131--21143, 2021.

\bibitem{zhang2024evaluation}
Lunwei Zhang, Siyuan Feng, Tiemin Li, and Yao Jiang.
\newblock \href{https://doi.org/10.1016/j.measurement.2024.114188}{Evaluation,
  selection and validation of force reconstruction models for vision-based
  tactile sensors}.
\newblock {\em Measurement}, 226:114188, 2024.

\bibitem{Wu2024Vision}
Tianyu Wu, Yujian Dong, Xiaobo Liu, Xudong Han, Yang Xiao, Jinqi Wei, Fang Wan,
  and Chaoyang Song.
\newblock \href{https://doi.org/10.1016/j.matdes.2024.112629}{Vision-based
  Tactile Intelligence with Soft Robotic Metamaterial}.
\newblock {\em Materials \& Design}, 238:112629, 2024.

\bibitem{Ma2018Probabilistic}
Qianli Ma, Zachariah Goh, Sipu Ruan, and Gregory~S Chirikjian.
\newblock \href{https://doi.org/10.1007/s10514-018-9744-3}{Probabilistic
  Approaches to the AXB = YCZ Calibration Problem in Multi-Robot Systems}.
\newblock {\em Autonomous Robots}, 42:1497--1520, 2018.

\bibitem{Muller2023WorldRobotics}
Christopher Müller.
\newblock
  \href{https://ifr.org/img/worldrobotics/Executive_Summary_WR_Industrial_Robots_2023.pdf}{World
  Robotics 2023 – Industrial Robots}.
\newblock Technical report, VDMA Services GmbH, 2023.

\bibitem{UR10e}
UR.
\newblock {\em
  \href{https://www.universal-robots.com/download/manuals-e-seriesur20ur30/user/ur10e/516/user-manual-ur10e-e-series-sw-516-english-international-en/}{User
  Manual - UR10e e-Series - Original Instructions (EN)}}.
\newblock Universal Robots, 10.4.186 edition, 2024.

\bibitem{UR5}
UR.
\newblock {\em
  \href{https://www.universal-robots.com/download/manuals-cb-series/user/ur5/315/user-manual-ur5-cb-series-sw315-english-international-en/}{User
  Manual - UR5/CB3 - Original Instructions (EN)}}.
\newblock Universal Robots, 9.3.116 edition, 2021.

\bibitem{FrankaEmika}
Franka.
\newblock {\em
  \href{https://download.franka.de/documents/100010_Product\%20Manual\%20Franka\%20Emika\%20Robot_10.21_EN.pdf}{Franka
  Emika Robot’s Instruction Handbook}}.
\newblock Franka, 10.21 edition, 2021.

\bibitem{AUBOi5}
AUBO.
\newblock {\em
  \href{https://www.aubo-cobot.com/public/assets/aubo/downloadsen/manualdl/AUBO-i5\%20\&\%20CB-M\%20User\%20Manual-V4.5.13-20230728.pdf}{AUBO
  - i5 \& CB - M User Manual}}.
\newblock AUBO, 4.5.13 edition, 2023.

\bibitem{Wong2018OptimalLinear}
Xue~Iuan Wong, Puneet Singla, Taewook Lee, and Manoranjan Majji.
\newblock \href{https://doi.org/10.1109/CVPRW.2018.00199}{Optimal Linear
  Attitude Estimator for Alignment of Point Clouds}.
\newblock In {\em IEEE/CVF Conference on Computer Vision and Pattern
  Recognition Workshops (CVPRW)}, pages 1577--15778. IEEE, 2018.

\bibitem{Rusu20113DIsHere}
Radu~Bogdan Rusu and Steve Cousins.
\newblock \href{https://doi.org/10.1109/ICRA.2011.5980567}{3D is Here: Point
  Cloud Library (PCL)}.
\newblock In {\em IEEE International Conference on Robotics and Automation
  (ICRA)}, pages 1--4. IEEE, 2011.

\bibitem{Raguram2012USAC}
Rahul Raguram, Ondrej Chum, Marc Pollefeys, Jiri Matas, and Jan-Michael Frahm.
\newblock \href{https://doi.org/10.1109/TPAMI.2012.257}{USAC: A Universal
  Framework for Random Sample Consensus}.
\newblock {\em IEEE Transactions on Pattern Analysis and Machine Intelligence},
  35(8):2022--2038, 2012.

\bibitem{Chen1992ObjectModelling}
Yang Chen and G{\'e}rard Medioni.
\newblock \href{https://doi.org/10.1109/ROBOT.1991.132043}{Object Modelling by
  Registration of Multiple Range Images}.
\newblock {\em Image and Vision Computing}, 10(3):145--155, 1992.

\bibitem{Mahadevan2021Intelligent}
Rishikesh Mahadevan, Avinaash Jagan, Lakshmi Pavithran, Ashutosh Shrivastava,
  and Senthil~Kumaran Selvaraj.
\newblock \href{https://doi.org/10.1016/j.matpr.2020.12.1149}{Intelligent
  Welding by using Machine Learning Techniques}.
\newblock {\em Materials Today: Proceedings}, 46:7402--7410, 2021.

\end{thebibliography}
\end{document}